\documentclass[a4paper,fleqn]{cas-dc}
\usepackage{tikz}
\usetikzlibrary{mindmap,trees,shapes.geometric}
\usepackage{longtable}
\usepackage{hyperref}
\usepackage{orcidlink}
\usetikzlibrary{positioning, arrows.meta, shapes, backgrounds}
\usepackage[x11names]{xcolor}
\usepackage[colorinlistoftodos,prependcaption,textsize=small]{todonotes}
\usepackage{eurosym}
\usepackage{xcolor} % Required to handle the colors
\usepackage{soul}   % Provides the \hl{} command
\usepackage[most]{tcolorbox}
\usepackage{listings}
\definecolor{MyBlue}{RGB}{26,13,171}
\definecolor{algorithmblue}{RGB}{35,105,175}

\lstdefinelanguage{json}{
    basicstyle=\ttfamily\small\color{black},
    string=[s]{"}{"},
    stringstyle=\color{black},
    comment=[l]{//},
    commentstyle=\color{gray},
    showstringspaces=false,
    breaklines=true,
    columns=fullflexible,
    keepspaces=true
}

\usepackage[numbers]{natbib}
\def\tsc#1{\csdef{#1}{\textsc{\lowercase{#1}}\xspace}}
\tsc{}
\tsc{}
\begin{document}
\let\WriteBookmarks\relax
\def\floatpagepagefraction{1}
\def\textpagefraction{.001}

\shorttitle{Explainable Plant Disease Diagnosis via Multi-Agent Fusion of CNNs and MLLMs}  

\shortauthors{R. Sapkota et al.}  

\title [mode = title]{Fusing Perceptual Vision Experts with Multimodal LLMs for Explainable Plant Disease Diagnosis: From Benchmark Imagery to Real-World Robotic Field Validation}

\tnotemark[1]

\tnotetext[1]{This work is supported by the National Science Foundation (NSF) and the United States Department of Agriculture (USDA), National Institute of Food and Agriculture (NIFA), through the "Artificial Intelligence (AI) Institute for Agriculture" program under Award Numbers AWD003473 and AWD004595, and USDA-NIFA Accession Number 1029004 for the project titled "Robotic Blossom Thinning with Soft Manipulators." Additional support was provided through USDA-NIFA Grant Number 2024-67022-41788, Accession Number 1031712, under the project "Expanding UCF AI Research To Novel Agricultural Engineering Applications (PARTNER)." }


%[<options>]

\author[1]{Ranjan Sapkota}[orcid=0000-0002-5417-6744]
\cormark[1]
\ead{rs2672@cornell.edu}
\affiliation[1]{organization={Cornell University, Department of Biological and Environmental Engineering},city={Ithaca},postcode={14850},state={NY},country={USA}}

\author[2,3]{Konstantinos I. Roumeliotis}[orcid=0000-0002-8098-1616]
\ead{k.roumeliotis@uop.gr}
\affiliation[2]{organization={University of the Peloponnese, Department of Informatics and Telecommunications},city={Tripoli},postcode={22131},country={Greece}}
\affiliation[3]{organization={Agricultural University of
Athens, Department of Agribusiness and Supply Chain Management},city={Athens},postcode={11855},country={Greece}}
\author[1]{Pengyao Xie}[orcid=add]
\ead{px62@cornell.edu}
\author[2]{Nikolaos D. Tselikas}[orcid=0000-0001-5799-3558]
\ead{ntsel@uop.gr}

\author[1]{Lirong Xiang}[orcid=0000-0003-1573-0906]
\ead{lxiang@cornell.edu}

\author[1]{Manoj Karkee}[orcid=0000-0001-5337-4848]
%\ead{mk2684@cornell.edu}
\cormark[1]

\cortext[1]{Corresponding authors}




\begin{abstract}
Reliable plant disease diagnosis requires fusing uncertain perceptual evidence into explainable and actionable decisions. We present a \textbf{Hybrid Hierarchical Multi-Agent Framework} (H\textsuperscript{2}MAF) combining decision-level fusion of EfficientNet-B3 and ConvNeXt-Tiny with semantic arbitration by the open-weight MLLMs Gemma~4~E4B and Qwen3.5~4B. Using images and structured JSON evidence, the framework generates diagnoses, arbitration rationales, risk levels, treatment urgency, and estimated financial exposure. Evaluation covers 14,364 images and 1,370 held-out test images from PlantDoc (2,922 images; 27 classes) and two non-public, robot-acquired Cornell field campaigns: Stage~2 (4,215 images; 20~GB) and Stage~4 (7,227 images; 40~GB). On PlantDoc, Gemma improves top-1 accuracy from 63.9\% to 68.5\%; within the 41.7\% CNN-disagreement subset, Qwen exceeds the stronger CNN by 7.6 percentage points. Cornell CNN accuracies range from 96.8\% to 99.8\%. MLLM accuracy reaches 99.3\% on Stage~2, slightly below the best CNN (99.8\%), and 98.9\% on Stage~4, exceeding the best CNN (98.3\%). Gemma's Critical-risk rate aligns closely with urgent-disease prevalence, whereas Qwen systematically over-assigns Critical risk by 3.5--14.4 points, highlighting the need for deployment-specific risk calibration. Code: \href{https://github.com/Applied-AI-Research-Lab/Explainable-AI-Plant-Disease-Detection}{GitHub}.
\end{abstract}


\maketitle
\scriptsize
% \tableofcontents
\normalsize
% Main text

%% ─────────────────────────────────────────────────────────────────────────────
\section{Introduction}
\label{sec:Introduction}

Plant diseases account for 20--40\% of global crop yield losses annually, posing a critical threat to fohighod security and rural economies \cite{Wolfert2017BigData}.
Early, accurate diagnosis is therefore essential, yet field deployment of disease detection systems remains constrained by three persistent gaps that the computer-vision and information-fusion literature has not fully bridged.

\textbf{Gap~1: Single-model perceptual fragility.}
Benchmark datasets such as PlantVillage \cite{Hughes2015PlantVillage,Mohanty2016PlantVillage} were captured under controlled laboratory conditions uniform backgrounds, consistent lighting, single leaves yielding top-1 accuracies exceeding 99\% for standard CNNs \cite{Ferentinos2018DL_plant}.
However, Barbedo \cite{Barbedo2018factors} and Singh et al. \cite{Singh_2020} demonstrated that the same models suffer drastic performance drops when applied to real-field imagery, where cluttered backgrounds, multiple overlapping leaves, and variable illumination prevail.
Any single classifier, however well trained, will therefore produce meaningfully different confidence distributions than an architecturally distinct classifier on the same challenging image, and neither is reliably ``more correct'' a priori: a principled way of \emph{fusing} their evidence is required rather than trusting either signal in isolation.

\textbf{Gap~2: The classification-to-action disconnect.}
Even a perfectly accurate classifier only outputs a class label.
A farm manager responsible for a \euro{}50,000 tomato crop needs to know \emph{how severe} the outbreak is, \emph{which treatment} to apply, \emph{how urgently}, and \emph{at what financial risk} if no action is taken.
Conventional pipelines offer no mechanism for this translation, leaving domain experts to interpret raw probability scores without decision support.

\textbf{Gap~3: The benchmark-to-real-world validation gap.}
The overwhelming majority of plant disease AI studies including our own prior work \cite{roumeliotis2025plantdiseasedetectionmultimodal} are validated exclusively on public, internet-curated, or laboratory-acquired benchmarks (PlantVillage, PlantDoc).
Such datasets, however carefully constructed, are static snapshots assembled by web scraping or single-session photography; they cannot fully capture the continuous, temporally correlated, sensor-noise-laden imagery produced by an actual autonomous field-monitoring platform operating on a working farm.
Recent work has confirmed that the controlled-to-field reliability gap persists even after standard mitigation techniques are applied \cite{Xiang2026Reliability}, yet almost no published study has evaluated a plant disease decision-support pipeline on \emph{genuine, non-public, robot-acquired} field data, because such data are rarely available outside the laboratories that collect them.

\textbf{Our contribution.}
This paper proposes a \textbf{Hybrid Hierarchical Multi-Agent Framework} (H\textsuperscript{2}MAF) that closes all three gaps through a principled two-stage information-fusion architecture and, critically validates that architecture on three progressively more realistic datasets including two previously unpublished, closed, field-acquired datasets collected by the authors.
The framework operates in three phases, implementing decision-level fusion followed by semantic-level fusion:

\begin{enumerate}
  \item \textbf{Perceptual Classification (decision-level fusion input).} Two architecturally distinct CNNs EfficientNet-B3 \cite{Tan2019EfficientNet} and ConvNeXt-Tiny \cite{Liu2022ConvNeXt} are independently fine-tuned on each dataset, producing probability distributions over the target disease classes.
  \item \textbf{Contextual Alignment.} CNN outputs are serialised alongside crop- or disease-specific business context (crop value, pathogen identity, weather conditions, risk-aversion level) into a structured JSON artefact that grounds MLLM reasoning and suppresses hallucination.
  \item \textbf{Cognitive Reasoning (semantic-level fusion).} Two small Multimodal LLMs, Google Gemma~4~E4B and Alibaba Qwen3.5~4B, receive the leaf image plus the JSON artifact and produce a structured Explainable AI (XAI) report containing a final diagnosis, arbitration reasoning, visual symptom description, risk level, business recommendation, and treatment window.
\end{enumerate}

This separation exploits the complementary strengths of each component: CNNs excel at fine-grained texture discrimination; MLLMs contribute commonsense agricultural knowledge, semantic conflict resolution, and natural-language explanation generation tasks for which they are inherently suited.
The framework is \emph{model-agnostic} and \emph{dataset-agnostic}: any fine-tuned CNN or instruction-following MLLM can be plugged in, and the same three-phase pipeline is applied, without architectural changes, to three datasets that differ radically in acquisition modality, class count, and visual difficulty.

The main contributions of this work are:
\begin{itemize}
  \item The first systematic study of MLLM-driven conflict arbitration between competing CNN perceptual experts validated not only on an internet-curated ``in-the-wild'' benchmark (PlantDoc) but also on two previously unpublished, continuously-captured, robot-acquired real-world field datasets (Cornell Stage~2, 20~GB; Cornell Stage~4, 40~GB), spanning 14,364 images and 1,370 held-out test images in total.
  \item Empirical demonstration, replicated across all three datasets, that MLLM arbitration value is a function of the vision-expert disagreement rate: substantial gains (+7.6~points) at a 41.7\% conflict rate (PlantDoc), and comparatively marginal gains at conflict rates of 1.7--4.1\% (Cornell Stage~2/4), where the CNN experts already agree on 95.9--98.3\% of images.
  \item A rigorous analysis of MLLM override behaviour across all three datasets, showing that models which override \emph{both} CNNs independently achieve only 0--17.6\% accuracy, establishing a design principle against unconstrained MLLM autonomy that generalises beyond a single dataset.
  \item A novel \emph{risk-personality calibration} analysis quantifying how closely each MLLM's assigned Critical-risk rate tracks the true prevalence of the most urgent disease class, replicated across the two Cornell datasets, revealing that Gemma is consistently well-calibrated (0.14--0.5-point error) while Qwen is consistently alarmist (3.5--14.4-point over-flagging).
  \item A fully reproducible, open-source pipeline integrating CNN training, JSON artefact generation, MLLM inference, and automated evaluation, applied uniformly across three independent datasets of increasing scale and realism.
\end{itemize}

The remainder of this paper is structured as follows.
Section~\ref{sec:RelatedWork} reviews the related literature on plant disease classification, multimodal LLMs, XAI, and information fusion for decision support.
Section~\ref{sec:Dataset} describes all three datasets: PlantDoc and the two Cornell real-world field datasets: and the experimental train/validation/test protocol for each.
Section~\ref{sec:Methodology} presents the three-phase H\textsuperscript{2}MAF architecture in detail.
Section~\ref{sec:Setup} defines the experimental setup and evaluation metrics.
Section~\ref{sec:Results} reports quantitative results on each of the three datasets individually, followed by a cross-dataset synthesis.
Section~\ref{sec:Discussion} interprets the findings, discusses limitations, and identifies future directions.
Section~\ref{sec:Conclusion} concludes the paper.

%% ─────────────────────────────────────────────────────────────────────────────
\section{Related Work}
\label{sec:RelatedWork}

\subsection{Plant Disease Classification}

Deep learning has dominated plant disease classification since Mohanty et al. \cite{Mohanty2016PlantVillage} demonstrated 99\% accuracy on PlantVillage using AlexNet.
Subsequent studies systematically benchmarked ResNet \cite{He2016ResNet}, DenseNet \cite{Huang2017DenseNet}, and EfficientNet \cite{Tan2019EfficientNet} variants, consistently achieving near-perfect accuracy on controlled images while confirming large performance drops on real-field data \cite{Barbedo2018factors,Kamilaris2018DL_agriculture}.
Singh et al. \cite{Singh_2020} introduced PlantDoc to address this gap; since its release, it has become the canonical benchmark for ``in-the-wild'' classification, with reported accuracies typically in the 60--75\% range for single CNNs.

ConvNeXt \cite{Liu2022ConvNeXt} represents a post-ViT modernisation of the convolutional paradigm, matching Vision Transformer \cite{dosovitskiy2020vit} performance while retaining CNN inductive biases crucial for small-data regimes.
We adopt ConvNeXt-Tiny as our second perceptual expert precisely because it outperforms Transformer-based alternatives on small training sets such as PlantDoc's 2,269-image training split.

\subsection{Multimodal LLMs in Visual Understanding}

The emergence of instruction-following Multimodal LLMs (MLLMs) from Flamingo \cite{Alayrac2022Flamingo} and GPT-4V \cite{achiam2023gpt4} to open-weight alternatives such as Gemma~4 \cite{Gemma4_2025} and Qwen3 \cite{Qwen3_2025} has introduced a new paradigm for visual reasoning.
Unlike CNNs, which learn discriminative features through supervised fine-tuning, MLLMs bring pre-trained commonsense, domain knowledge, and language generation capabilities that enable rich, context-aware explanations \cite{Lu2023MLLMsurvey}.

Roumeliotis et al. \cite{roumeliotis2025plantdiseasedetectionmultimodal} compared GPT-4 fine-tuned for plant disease classification against ResNet-50, finding that LLMs can encode domain knowledge competitive with supervised CNNs.
However, direct MLLM classification of fine-grained disease textures without any CNN assistance remains unreliable at the confidence levels required for actionable recommendations.
Our work differs fundamentally by \emph{not} replacing CNNs with MLLMs but instead positioning MLLMs as arbitrators that consume pre-computed CNN signals a form of hierarchical decision fusion in which each modality contributes at the layer where it is strongest.
We also note that none of these prior studies including our own were validated beyond internet-scraped or laboratory-curated imagery; the present work is, to our knowledge, the first to test this fusion paradigm against genuine, closed, robot-acquired field data.

\subsection{Explainable AI in Agriculture}

XAI methods such as LIME \cite{Ribeiro2016LIME}, SHAP \cite{Lundberg2017SHAP}, and Grad-CAM \cite{Selvaraju2017GradCAM} have been applied to visualise CNN decisions in plant pathology \cite{Arrieta2020XAI}, but they produce pixel-level saliency maps that require agronomist expertise to interpret.
No existing work translates CNN classifications into structured, actionable crop management recommendations with quantified financial impact the primary novelty of our XAI layer.

\subsection{Transfer Learning and Domain Adaptation}

Transfer learning from ImageNet pre-trained weights has become the de facto training protocol for plant disease classification \cite{Kamilaris2018DL_agriculture}.
Mohanty et al. \cite{Mohanty2016PlantVillage} demonstrated that even shallow fine-tuning of VGG-16 and AlexNet on PlantVillage yields 97--99\% accuracy, establishing the paradigm.
Subsequent work explored fine-tuning depth: Ferentinos \cite{Ferentinos2018DL_plant} compared VGG-16, AlexNet, GoogLeNet, and a custom architecture, finding that deeper unfreeze strategies and higher-resolution inputs consistently improve generalisation on in-the-wild samples.

However, transfer from ImageNet to agricultural images presents a domain gap: ImageNet features are tuned for object recognition (edges, shapes, object parts), while disease detection requires sensitivity to subtle textural patterns (concentric ring formation, sporulation density, lesion colouration gradients) that are not well-represented in ImageNet classes.
Partial-unfreeze strategies \cite{Tan2019EfficientNet} balance preservation of low-level texture detectors with task-specific adaptation of mid- and high-level features.
In the present work, we apply partial unfreeze (top 30\%) for EfficientNet-B3 and full fine-tune for ConvNeXt-Tiny \cite{Liu2022ConvNeXt} uniformly across all three datasets, the latter being justified by ConvNeXt's stronger per-layer feature reuse from its inverted bottleneck design.

\subsection{Ensemble Methods and Multi-Expert Model Fusion}

Ensembling multiple models to improve robustness has a long history in computer vision, from classical bagging and boosting to modern deep ensemble and mixture-of-experts architectures.
In the plant disease domain, ensemble approaches combining CNNs with different architectures or training regimes consistently outperform single-model baselines by 3--8 percentage points on challenging benchmarks \cite{Kamilaris2018DL_agriculture}.
However, standard ensembles aggregate predictions through fixed rules (majority vote, average probability) without any semantic understanding of why the models disagree or what the disagreement implies operationally.
More recently, Roumeliotis et al. \cite{roumeliotis11373381} proposed a modular agentic AI framework with trust-aware orchestration and retrieval-augmented reasoning for visual classification, demonstrating that orchestrator-agent trust mechanisms can improve calibration and reduce overconfident misclassifications---a direction complementary to the semantic arbitration approach adopted in this work.

Our framework differs from traditional ensembling in a fundamental way: the second-stage arbitration is performed by a model (MLLM) that has not been trained on the target dataset and has no direct access to class probability statistics.

The MLLM introduces a new, orthogonal information source pre-trained visual-semantic knowledge that is architecturally decoupled from the first-stage CNN uncertainty.
This is closer in spirit to a human expert reviewing conflicting laboratory reports: the expert does not simply average the reports, but applies domain knowledge to determine which is more credible given the available visual evidence.
This work extends that principle by testing whether the same fusion logic transfers unchanged from a 27-class, internet-curated benchmark to a 3-class, continuously-captured, real-world robotic field dataset a substantially different information-fusion regime in which the CNN experts rarely disagree at all.

\subsection{Prompt Engineering for Structured Output}

Instructing LLMs to produce structured output (JSON, XML, SQL) is an active research area.
Naive prompting frequently results in verbose free-form responses that are difficult to parse reliably.
Key techniques include explicit format specification \cite{achiam2023gpt4}, chain-of-thought suppression for latency-sensitive applications \cite{Qwen3_2025}, and constrained decoding approaches that enforce syntactic validity.

Across all three datasets used in this study, we found a consistent distinction between the two MLLM families: Gemma~4~E4B follows JSON formatting instructions reliably in zero-shot mode, while Qwen3.5~4B's built-in chain-of-thought reasoning mode consumes the generation budget with verbose reasoning before producing structured output.
Suppressing thinking mode via \texttt{enable\_thinking=False} in the chat template resolved this issue, yielding coverage above 96.8\% on every dataset.
This finding has broader implications for deploying reasoning-capable MLLMs in production pipelines where structured output is mandatory.

Power \cite{Power2003DSS} surveyed the evolution of Decision Support Systems (DSS) from rule-based expert systems to data-driven models.
Wolfert et al. \cite{Wolfert2017BigData} identified smart-farming data integration as a key challenge for modern agricultural DSS.
Our framework takes a first step toward addressing this challenge by embedding crop- and disease-specific economic context including estimated crop value, pathogen identity, and a risk-aversion level into the structured JSON artefact consumed by the MLLM, enabling the reasoning layer to generate outputs that go beyond bare classification labels to include treatment urgency, risk level, and estimated financial impact, consistently across benchmark and real-world data.

%% ─────────────────────────────────────────────────────────────────────────────
\section{Datasets: From Internet Benchmark to Real-World Robotic Field Validation}
\label{sec:Dataset}

\begin{figure*}[htb]
\centering
\includegraphics[width=0.95\linewidth]{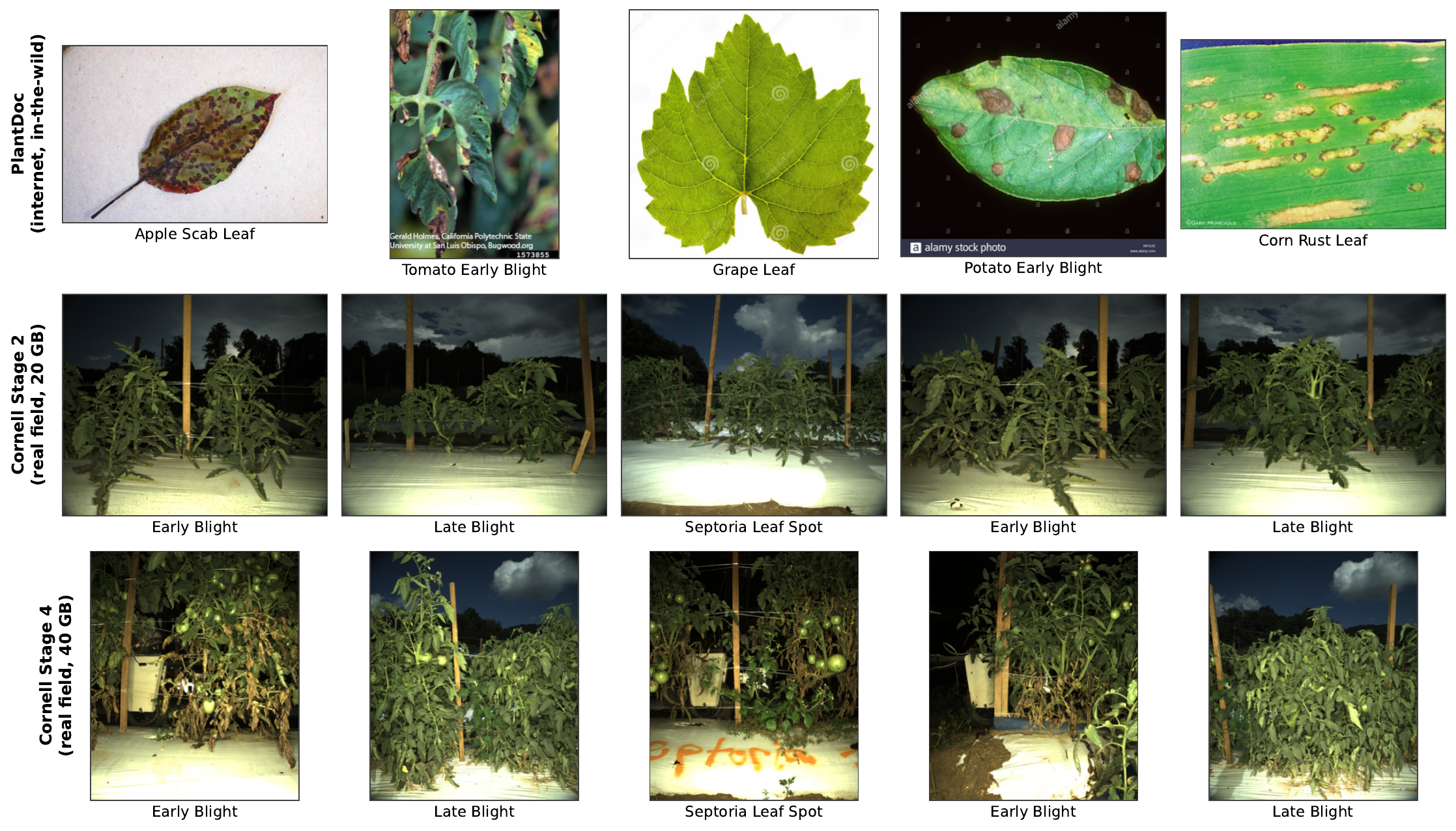}
\caption{Representative sample images from the three datasets used in this study. Top row: PlantDoc, internet-scraped ``in-the-wild'' photographs (27 classes). Middle and bottom rows: Cornell Stage~2 and Stage~4, continuously-captured, autonomous field-robot imagery (3 classes). The visual contrast between crowd-sourced internet photography and continuous field-video frames illustrates the acquisition-modality gap discussed throughout this section.}
\label{fig:dataset_samples}
\end{figure*}

A central methodological contribution of this study is the validation of the same H\textsuperscript{2}MAF pipeline on three datasets of markedly different provenance, scale, and realism.
% Table~\ref{tab:dataset_overview} summarises the three datasets; the remainder of this section describes each in turn.
Table~\ref{tab:dataset_overview} summarises the three datasets; Fig.~\ref{fig:dataset_samples} provides representative sample images from each; the remainder of this section describes each in turn.

\begin{table*}[htb]
\centering
\caption{Overview of the three datasets used in this study. ``Acquisition'' describes how the images were captured, whereas ``Setting'' indicates whether the source is a public Internet-curated benchmark or a closed, real-world field dataset shared under a research collaboration.}
\label{tab:dataset_overview}
\scriptsize
\setlength{\tabcolsep}{4pt}
\renewcommand{\arraystretch}{1.15}
\begin{tabular}{lccccl}
\hline
\textbf{Dataset} &
\textbf{Images} &
\textbf{Classes} &
\textbf{Test $N$} &
\textbf{Acquisition} &
\textbf{Setting} \\
\hline
PlantDoc \cite{Singh_2020,GuerreroSerna2025PlantDoc}
& 2,922 & 27 & 252
& Internet-scraped photographs
& Public, in-the-wild benchmark \\

Cornell Stage~2 (this work)
& 4,215 & 3 & 403
& Autonomous field-robot video frames, 20~GB
& Closed, real-world field data \\

Cornell Stage~4 (this work)
& 7,227 & 3 & 715
& Autonomous field-robot video frames, 40~GB
& Closed, real-world field data \\
\hline
\textbf{Total}
& \textbf{14,364}
& ---
& \textbf{1,370}
& ---
& --- \\
\hline
\end{tabular}
\end{table*}

\subsection{PlantDoc: An Internet-Curated ``In-the-Wild'' Benchmark}
\label{sec:dataset_plantdoc}

PlantDoc \cite{Singh_2020} is the primary public benchmark for ``in-the-wild'' plant disease classification.
Unlike PlantVillage \cite{Hughes2015PlantVillage}, whose images were captured under controlled laboratory conditions, PlantDoc consists of images scraped from the internet, reflecting the visual diversity of casual field photography: cluttered multi-leaf backgrounds, variable illumination, mixed healthy/diseased content within a single frame, and significant compression artefacts.

The Kaggle Version~7 of PlantDoc \cite{GuerreroSerna2025PlantDoc} used in this study comprises \textbf{2,922 images} organised into a predefined train/test split: 2,670 training images across 27 class folders and 252 test images across 26 class folders (one class, \texttt{Tomato\_two\_spotted\_spider\_mites\_leaf}, has only 2 training images and is absent from the test set, and is therefore excluded from all experiments).
The final dataset used for training and evaluation covers \textbf{27 classes} spanning 13 plant species, including both diseased and healthy leaf categories.
Table~\ref{tab:dataset} summarises the per-class distribution; significant imbalance is present (\texttt{Tomato\_leaf\_yellow\_virus}: 223 train images vs.\ \texttt{Bell\_pepper\_leaf}: 34, a 6.6$\times$ ratio), necessitating Focal Loss \cite{Lin2017FocalLoss} during CNN training.

A stratified 85/15 split of the 2,670 training images creates a validation set for model selection: 2,269 images for training and 401 for validation.
The 252 test images from the original PlantDoc split are held out entirely and used only for final evaluation.

\begin{table}[htb]
\centering
\caption{PlantDoc class distribution (27 classes used in experiments).}
\label{tab:dataset}
\scriptsize
\begin{tabular}{lcc}
\hline
\textbf{Class} & \textbf{Train} & \textbf{Test} \\
\hline
Apple Scab Leaf          & 83  & 10 \\
Apple Leaf               & 79  & 9  \\
Apple Rust Leaf          & 96  & 10 \\
Bell Pepper Leaf         & 34  & 8  \\
Bell Pepper Leaf Spot    & 74  & 9  \\
Blueberry Leaf           & 106 & 11 \\
Cherry Leaf              & 47  & 10 \\
Corn Gray Leaf Spot      & 63  & 4  \\
Corn Leaf Blight         & 182 & 12 \\
Corn Rust Leaf           & 107 & 10 \\
Peach Leaf               & 103 & 9  \\
Potato Early Blight      & 157 & 14 \\
Potato Late Blight       & 200 & 8  \\
Raspberry Leaf           & 112 & 7  \\
Soyabean Leaf            & 57  & 8  \\
Squash Powdery Mildew    & 124 & 6  \\
Strawberry Leaf          & 88  & 8  \\
Tomato Early Blight      & 79  & 9  \\
Tomato Septoria Leaf Spot& 145 & 12 \\
Tomato Leaf              & 44  & 8  \\
Tomato Bacterial Spot    & 101 & 9  \\
Tomato Late Blight       & 101 & 10 \\
Tomato Mosaic Virus      & 44  & 10 \\
Tomato Yellow Virus      & 223 & 15 \\
Tomato Mold              & 85  & 6  \\
Grape Leaf               & 63  & 12 \\
Grape Black Rot          & 71  & 8  \\
\hline
\textbf{Total}           & \textbf{2,670} & \textbf{252} \\
\hline
\end{tabular}
\end{table}

\subsection{Cornell Real-World Field Datasets: Stage~2 and Stage~4}
\label{sec:dataset_cornell}

To address the benchmark-to-real-world validation gap identified in Section~\ref{sec:Introduction}, this study incorporated two customized, non-public datasets acquired by the co-authors using a robotic field-phenotyping platform. In contrast to PlantDoc, which primarily comprises independently sourced Internet images captured with heterogeneous backgrounds, viewpoints, and imaging devices, the Cornell datasets were generated through systematic robotic imaging of tomato canopies under outdoor field conditions. The datasets therefore represent the type of continuous visual information that would be encountered by an autonomous crop-monitoring platform during field deployment.

The broader field trials were established at two research locations in North Carolina, USA: the Mountain Research Station (MRS) in Waynesville and the Mountain Horticultural Crops Research and Extension Center (MHCREC) in Mills River. The three disease classes evaluated in the present study were obtained from disease-specific experimental plots at MRS. Late Blight imagery originated from Field~A13, whereas Early Blight and Septoria Leaf Spot imagery originated from Field~C22. Maintaining disease-specific plots provided a direct connection between the acquired images and the corresponding field treatments, inoculation records, and disease-monitoring protocols.

The study focused on three economically important foliar diseases of tomato: \textbf{Early Blight} (\textit{Alternaria solani}), \textbf{Late Blight} (\textit{Phytophthora infestans}), and \textbf{Septoria Leaf Spot} (\textit{Septoria lycopersici}). These diseases were selected because they can exhibit partially overlapping visual symptoms, including chlorotic regions, necrotic lesions, and progressive foliar deterioration, while differing substantially in epidemiology, progression rate, treatment urgency, and potential economic impact. Their visual similarities make reliable discrimination particularly important for automated field diagnosis and downstream crop-management decisions.

The field trials included both determinate and indeterminate tomato cultivars, thereby introducing variation in plant morphology, canopy density, growth habit, and symptom visibility. Plants were established in single rows with an intra-row spacing of 0.457~m and a row-center spacing of 1.524~m. White plastic mulch and drip irrigation were used throughout the experimental plots. These production conditions created realistic sources of visual complexity, including overlapping foliage, nonuniform canopy structure, mulch reflections, soil and background interference, and partial occlusion of symptomatic leaves.

The Early Blight and Septoria Leaf Spot plots were inoculated within two weeks of planting to promote disease establishment under the corresponding experimental treatments. In contrast, no artificial inoculation was applied in the Late Blight plot. Disease identity was determined from the disease-specific experimental design together with the associated plot assignments, inoculation procedures, and treatment records. Disease development was additionally monitored through repeated visual assessments of individually marked plants using a modified Horsfall--Barratt disease-severity scale. This combination of controlled plot organization, experimental records, and repeated in-field assessments provided the biological and agronomic foundation for assigning disease labels to the robot-acquired images used in the present study.

From the broader four-stage field-acquisition sequence, Stage~2 and Stage~4 were selected as two temporally separated datasets representing different periods of canopy and disease development. Their inclusion enabled the proposed information-fusion framework to be evaluated using earlier- and later-season robotic imagery rather than relying exclusively on a single acquisition period or an Internet-curated benchmark. The robotic imaging configuration, temporal acquisition protocol, and subsequent image-screening and dataset-organization procedures are described in the following subsections.

\subsubsection{Robotic Image-Acquisition Platform}
\label{sec:robotic_acquisition_platform}

\begin{figure*}[htbp]
    \centering
    \includegraphics[width=0.98\textwidth]{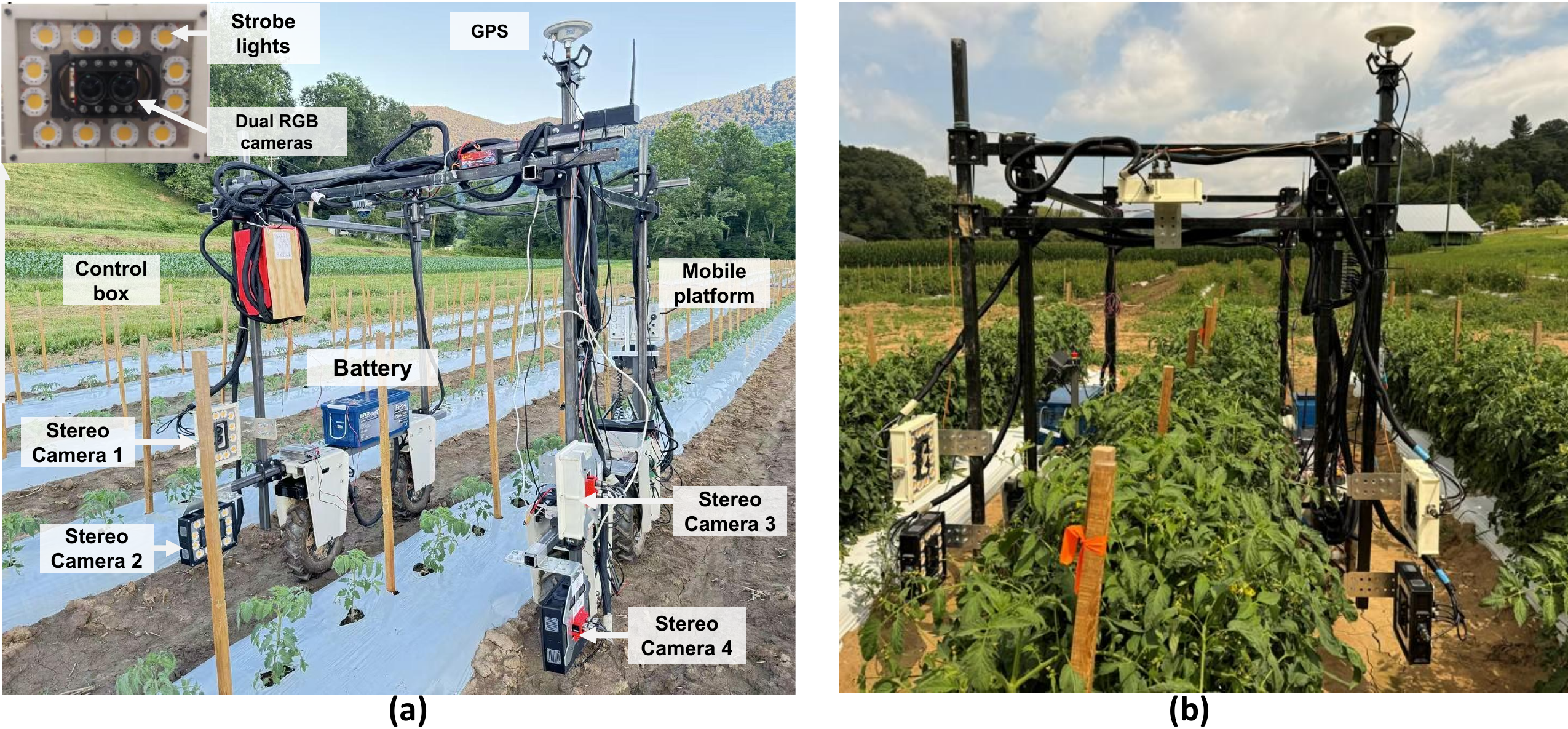}
    \caption{Demonstration of the customized Amiga robotic phenotyping platform used to collect the Cornell tomato-disease datasets under outdoor field conditions: (a) the robotic data-collection prototype operating during the earlier-season campaign, and (b) the same prototype operating within the denser tomato canopy during the later-season campaign. The platform incorporated five PhenoStereo imaging units, comprising ten RGB cameras and synchronized active strobe-lighting assemblies, to acquire multi-view canopy images while navigating along the crop rows.}
    \label{fig:cornell_robot_platform}
\end{figure*}

Image acquisition was performed using a customized robotic phenotyping system developed by integrating an Amiga agricultural mobile robot with five PhenoStereo imaging units. The individual PhenoStereo units were configured according to the imaging principles reported by He et al.~\cite{he2025high}, while their five-unit arrangement, structural integration with the Amiga platform, and multi-view field-acquisition strategy were developed for the tomato disease campaigns investigated in this study. Figure~\ref{fig:cornell_robot_platform} illustrates the physical configuration and operation of the same robotic platform at two different stages of the tomato-growing season. These photographs demonstrate how the data-collection prototype operated under contrasting earlier- and later-season canopy conditions. The platform was designed to navigate along tomato crop rows while continuously acquiring canopy images from complementary positions, heights, and viewing directions. The five imaging units were spatially distributed across the robotic structure to observe different vertical and lateral portions of the canopy during a single traversal. Images were collected from both sides of the crop rows to increase spatial coverage and capture symptomatic foliage that could remain occluded when viewed from only one direction.

Each PhenoStereo unit incorporated two Blackfly S USB3 machine-vision cameras (FLIR Systems Inc., Wilsonville, OR, USA), resulting in a total of ten RGB cameras across the complete robotic platform. Each camera was equipped with a 1/1.8-in.\ complementary metal--oxide--semiconductor (CMOS) image sensor and a 4-mm focal-length lens and recorded images at a native resolution of $2448 \times 2048$ pixels. Within each imaging unit, the two cameras were separated by a 38-mm stereo baseline, enabling binocular observation at the relatively short working distances encountered between the robotic platform and tomato canopies. The optical configuration provided approximate diagonal, horizontal, and vertical fields of view of $94.0^{\circ}$, $82.9^{\circ}$, and $66.5^{\circ}$, respectively. Both the left and right RGB images generated by each stereo-camera pair were saved during acquisition. Although the PhenoStereo configuration supported stereoscopic imaging and could provide depth information, only the acquired RGB images were used in the disease-classification experiments presented in this study. Stereo-derived depth information was not supplied as an input to either the CNN perceptual experts or the MLLM reasoning components.

To support reliable image acquisition under variable outdoor illumination, each PhenoStereo unit was surrounded by 12 high-intensity flashlights (Bridgelux Inc., Fremont, CA, USA) positioned around the corresponding stereo-camera pair. The camera exposures and strobe-lighting system were electronically synchronized to permit short exposure times and reduce image degradation caused by robotic-platform vibration, wind-induced leaf motion, and temporal changes in ambient irradiance. The illumination assemblies were arranged so that the actively illuminated canopy region approximately coincided with the field of view of the corresponding stereo cameras. This configuration reduced the influence of direct sunlight, backlighting, and strong canopy shadows, thereby improving the visibility and consistency of foliar symptoms across cameras and acquisition periods.

The PhenoStereo hardware supported acquisition at rates of up to 14 stereoscopic image pairs per second. During the field campaigns, image capture was triggered automatically and operated continuously while the Amiga robot navigated along the tomato crop rows. The left- and right-camera images from all five imaging units were stored for subsequent quality screening, labeling, and model processing. The combination of high acquisition frequency and distributed camera placement enabled the platform to record dense image sequences from multiple perspectives as it passed the crop canopy. Nevertheless, synchronized active illumination did not eliminate all sources of field variability. The collected images retained natural variations associated with plant architecture, canopy density, camera-to-canopy distance, viewing angle, leaf orientation, foliage overlap, canopy occlusion, wind-induced motion, and residual environmental illumination. Consequently, the resulting datasets represent high-resolution, robot-acquired canopy imagery collected under realistic outdoor field conditions rather than isolated-leaf photographs acquired within a controlled laboratory environment.

\subsubsection{Temporal Acquisition Protocol}
\label{sec:temporal_acquisition_protocol}

Field imaging was conducted at four temporal stages between July and August 2025 to capture changes in tomato-canopy development and disease expression over the growing season. Stage~1 was acquired on July 2--3, Stage~2 on July 16--18, Stage~3 from July 30 to August 1, and Stage~4 from August 15--17, 2025. Together, these campaigns formed a four-stage field-imaging sequence spanning the principal period of disease development. The present study uses Stage~2 and Stage~4 because they represent two temporally separated observations of disease and canopy development and provide sufficiently large image collections for model training, validation, and testing. Stage~1 and Stage~3 were not included in the current experiments.

Stage~4 represents an independently conducted, later-season acquisition campaign rather than a strict numerical or image-level superset of Stage~2. Maintaining these campaigns as separate datasets enabled the framework to be evaluated using imagery collected at different stages of canopy and disease development. Each retained image preserved acquisition-provenance information derived from its collection stage, acquisition date, and timestamp. The robotic acquisitions were also linked to the geospatial context of the mapped disease plots, supporting traceability between the images, their corresponding field experiments, and the associated disease treatments.

The raw image files contained timestamps with millisecond precision. These timestamps confirmed that images within each acquisition sequence were captured in rapid, near-continuous succession, typically at intervals of approximately 0.3--1.0~s. Consequently, adjacent frames may depict the same plant, leaf, or symptomatic region from slightly different robotic-platform positions. The total number of captured frames was therefore substantially greater than the number of biologically independent plants or symptomatic regions represented in the datasets.

This sequential acquisition structure differs fundamentally from the independently sourced photographs contained in PlantDoc and reflects the continuous visual data stream generated by a mobile field-imaging platform. However, it also introduces temporal correlation among neighboring frames. This dependence must be considered when interpreting frame-level classification performance because visually similar observations of the same plant or lesion may occur within the collected image sequence.

\subsubsection{Image Screening, Annotation, and Dataset Organization}
\label{sec:cornell_dataset_organization}

Following acquisition, the raw image collections underwent a documented, multi-step screening and curation procedure conducted by the Cornell Automation and Robotics Laboratory. First, the frames were manually inspected to distinguish images containing genuine diseased-leaf or canopy content from images dominated by soil and unrelated background regions. Second, frames in which soil or background content had been incorrectly identified as diseased tissue were corrected to the appropriate healthy or background designation during source-data curation and excluded when they did not belong to the final three-disease classification dataset. Third, blank, completely dark, corrupted, or otherwise unusable frames were discarded. Finally, only RGB imagery was retained for the experiments reported in this study.

Image-level disease labels were derived from the identities of the disease-specific field plots and their corresponding inoculation and treatment records. The same individually marked plants were visually assessed repeatedly throughout the growing season using a modified Horsfall-Barratt disease-severity scale. The preliminary image labels and visual content were subsequently reviewed through the multi-step expert curation process. Images that did not contain interpretable visual evidence consistent with their assigned disease category were corrected or excluded from the final dataset. This combination of plot-level experimental records, repeated plant assessment, and post-acquisition image review provided additional label-quality control beyond reliance on filename or folder identity alone.

After quality control, the Stage~2 dataset contained 4,215 images distributed among 1,533 Early Blight, 1,478 Late Blight, and 1,204 Septoria Leaf Spot images. The Stage~4 dataset contained 7,227 images, including 3,295 Early Blight, 1,520 Late Blight, and 2,412 Septoria Leaf Spot images. The two datasets were maintained separately to preserve their temporal acquisition identities and enable evaluation across earlier- and later-season field campaigns.

Within each stage-specific dataset, every retained image was assigned to one of the three disease classes. Acquisition dates and timestamps were preserved to maintain temporal provenance, and the available geospatial and treatment records supported traceability to the corresponding experimental plots. Automated image-integrity checks were additionally applied during data loading to detect corrupted or unreadable files. Dataset partitioning was completed before model-specific resizing and augmentation, ensuring that all processed or augmented versions derived from a given image remained within the same data subset.

For the experiments reported in this study, each stage-specific dataset was divided using a stratified, frame-level random partition comprising 80\% training data, 10\% validation data, and 10\% test data, with a fixed random seed of 42. Stratification preserved the relative representation of Early Blight, Late Blight, and Septoria Leaf Spot across the three partitions. Because the source data were acquired as temporally successive frames, neighboring images may remain visually correlated across the randomly generated subsets. The resulting performance should therefore be interpreted as frame-level classification under the present partitioning protocol rather than as fully independent plant-, row-, or acquisition-session-level generalization.

Overall, the Cornell datasets complement PlantDoc by introducing images generated by a mobile robotic platform operating within natural tomato canopies. They capture field-specific variability associated with canopy occlusion, plant complexity, changing viewpoints, nonuniform symptom scale, wind-induced movement, and residual illumination variation that is not consistently represented in Internet-curated leaf-image benchmarks. Their labels additionally benefit from disease-specific plot identities, inoculation and treatment records, repeated visual assessment of marked plants, and expert image screening. Nevertheless, differences in class count, crop-host range, acquisition modality, and temporal dependence mean that absolute performance values should not be compared directly with PlantDoc without appropriate qualification.

\subsubsection{Cornell Stage~2 (20~GB)}

The Stage~2 release comprises \textbf{4,215 images} across the three disease classes: Early Blight (1,533), Late Blight (1,478), and Septoria Leaf Spot (1,204).
No predefined train/test split accompanies the raw data, so we construct a stratified 80/10/10 train/validation/test split (random seed 42) for reproducibility, consistent with the PlantDoc protocol's use of a held-out test partition.
After the automated pipeline (Section~\ref{sec:Methodology}) filters a small number of corrupted or unreadable frames, the evaluated test partition comprises \textbf{403 images}: 147 Early Blight, 141 Late Blight, and 115 Septoria Leaf Spot.

\subsubsection{Cornell Stage~4 (40~GB)}

The Stage~4 release is a substantially larger, independently acquired follow-up campaign comprising \textbf{7,227 images}: Early Blight (3,295), Late Blight (1,520), and Septoria Leaf Spot (2,412)-nearly double the total volume of Stage~2, with the largest proportional increase in the Early Blight and Septoria classes.
The same stratified 80/10/10 split protocol (seed 42) yields a nominal test partition of 723 images, of which \textbf{715} are successfully processed end-to-end by the pipeline (329 Early Blight, 146 Late Blight, 240 Septoria Leaf Spot); the eight excluded frames failed automated image-integrity checks and were skipped by the data loader rather than mis-imputed.

\subsubsection{Business Context for the Cornell Diseases}

Consistent with the PlantDoc business-context module (Section~\ref{sec:Methodology}), each Cornell disease class is mapped to a disease-specific business-intelligence record containing the causal pathogen, an estimated crop value at risk (\euro{}35,000 for Early and Late Blight, \euro{}50,000 for Septoria Leaf Spot, reflecting typical tomato field values), a risk-aversion level, and a treatment-urgency label.
Late Blight is the only class flagged \emph{high} risk-aversion / \emph{urgent} treatment-urgency in this lookup table, reflecting its well-documented capacity for rapid, epidemic-scale crop destruction (\textit{Phytophthora infestans} was the causal agent of the Irish Potato Famine); this single design choice becomes analytically important in Section~\ref{sec:Results}, where it provides a ground-truth reference against which each MLLM's risk-assignment calibration can be quantitatively measured.

\subsection{Cross-Dataset Comparison and Rationale}

Table~\ref{tab:dataset_comparison} contextualises all three datasets within the broader landscape of plant disease image datasets, and Table~\ref{tab:dataset_overview} (above) summarises their scale.
PlantVillage's controlled acquisition protocol makes it unsuitable for evaluating robustness to real-world conditions; we therefore do not use it directly, but cite its reported accuracy ceiling as context.
PlantDoc's substantially lower CNN accuracy (60--75\% in the literature) directly reflects the domain gap between laboratory and internet-sourced field imagery, and we retain it here precisely because it provides a challenging, conflict-rich classification scenario (41.7\% CNN disagreement rate; Section~\ref{sec:Results}) that motivates and stress-tests the MLLM arbitration layer.
The two Cornell datasets complement PlantDoc by offering something no public benchmark can: genuine, continuously-captured, non-curated field imagery from an operational research platform, at two different scales (20~GB and 40~GB), enabling us to test whether the H\textsuperscript{2}MAF fusion architecture---and its associated findings regarding conflict-dependent MLLM utility and risk-personality calibration---generalise beyond a single dataset and a single acquisition modality.

\begin{table*}[htb]
\centering
\caption{Comparison of plant disease image datasets relevant to this study.}
\label{tab:dataset_comparison}
\scriptsize
\begin{tabular}{lcccc}
\hline
\textbf{Dataset} & \textbf{Images} & \textbf{Classes} & \textbf{Setting} & \textbf{CNN Acc.} \\
\hline
PlantVillage \cite{Hughes2015PlantVillage} & 54,306 & 38 & Lab, public     & $>$99\% \\
PlantDoc \cite{Singh_2020}                &  2,598 & 27 & Field, public   & 60--75\% \\
PlantDoc v7 (this work)                  &  2,922 & 27 & Field, public   & 63.9\% \\
Cornell Stage~2 (this work)              &  4,215 &  3 & Field, closed   & 99.8\% \\
Cornell Stage~4 (this work)              &  7,227 &  3 & Field, closed   & 98.3\% \\
\hline
\end{tabular}
\end{table*}

We note an important caveat regarding the Cornell datasets' near-ceiling CNN accuracy (96--99.8\%; Section~\ref{sec:Results}), which is far higher than PlantDoc's 59--64\%.
Three factors jointly explain this gap: first, the Cornell task is genuinely easier in a taxonomic sense (3 visually distinct diseases on a narrow host range vs.\ 27 classes across 13 species with substantial inter-class visual similarity); second, because images are drawn from continuous video sequences, the stratified random split may place temporally adjacent (and hence highly similar) frames of the same physical lesion into different partitions, which likely inflates measured accuracy relative to a deployment scenario in which the system encounters a truly novel plant; and third, the Cornell images are full plant-canopy views captured at a consistent distance by the robot platform, whereas PlantDoc images are individual leaf photographs with highly variable framing, background clutter, and scale---a difference in image format that further contributes to the difficulty gap.
We return to this limitation in Section~\ref{sec:Discussion} and recommend session-level (rather than frame-level) splitting and localised lesion-patch cropping for future work using this data modality.

%% ─────────────────────────────────────────────────────────────────────────────
\section{Methodology}
\label{sec:Methodology}

\subsection{Framework Overview}

\begin{figure*}[htb]
\centering
\begin{tikzpicture}[
  node distance=0.5cm and 0.4cm,
  box/.style={rectangle, draw=MyBlue, thick, rounded corners=3pt,
              fill=blue!6, minimum width=3.0cm, minimum height=0.65cm,
              font=\scriptsize\sffamily, align=center},
  smallbox/.style={rectangle, draw=gray!70, thick, rounded corners=2pt,
                   fill=gray!10, minimum width=2.5cm, minimum height=0.55cm,
                   font=\scriptsize\sffamily, align=center},
  arr/.style={-Stealth, thick, MyBlue},
  garr/.style={-Stealth, thick, gray!60}
]
  \node[box, fill=green!10, draw=green!60!black]  (img)   {Leaf Image\\(PlantDoc / Cornell\\Stage 2 / Stage 4)};

  \node[box, right=0.7cm of img, yshift= 0.6cm]  (eff)   {EfficientNet-B3\\(Phase 1)};
  \node[box, right=0.7cm of img, yshift=-0.6cm]  (cnx)   {ConvNeXt-Tiny\\(Phase 1)};

  \node[box, right=0.7cm of eff, yshift=-0.6cm]  (align) {JSON Artefact\\(Phase 2)};
  \node[smallbox, below=0.4cm of align]           (ctx)   {Business Context};

  \node[box, right=0.8cm of align, yshift= 0.5cm] (gem)  {Gemma 4 E4B\\(Phase 3)};
  \node[box, right=0.8cm of align, yshift=-0.5cm] (qwn)  {Qwen3.5 4B\\(Phase 3)};

  \node[box, fill=orange!10, draw=orange!70, right=0.7cm of gem, yshift=-0.5cm] (xai) {XAI Report\\+ Decision};

  \draw[arr] (img.north east) -- (eff.west);
  \draw[arr] (img.south east) -- (cnx.west);
  \draw[arr] (eff.east) -- (align.north west);
  \draw[arr] (cnx.east) -- (align.south west);
  \draw[garr] (ctx.north) -- (align.south);
  \draw[arr] (align.east) -- (gem.west);
  \draw[arr] (align.east) -- (qwn.west);
  \draw[arr] (gem.east) -- (xai.north west);
  \draw[arr] (qwn.east) -- (xai.south west);
\end{tikzpicture}
\caption{H\textsuperscript{2}MAF pipeline, applied uniformly across the three datasets. Two CNN experts (Phase~1) feed a structured JSON artefact (Phase~2) which, together with the original image, is processed by two MLLM reasoning agents (Phase~3) to produce structured XAI reports.}
\label{fig:framework}
\end{figure*}
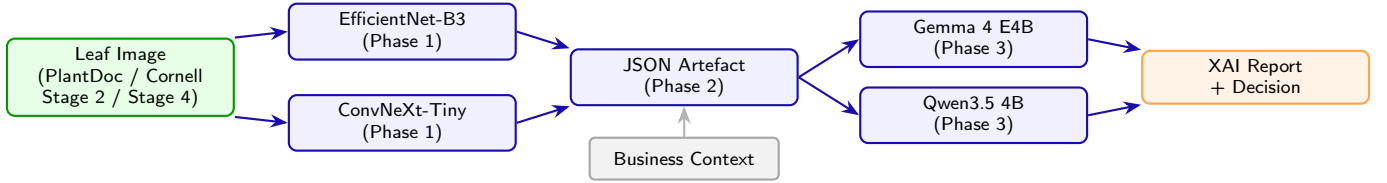

Fig.~\ref{fig:framework} illustrates the H\textsuperscript{2}MAF pipeline, which is applied identically to all three datasets described in Section~\ref{sec:Dataset}.
A leaf image enters Phase~1, where two CNN perceptual experts independently produce top-$K$ probability distributions over the dataset's target classes (27 for PlantDoc; 3 for each Cornell dataset).
Phase~2 assembles these predictions alongside dataset-specific business context into a structured JSON artefact.
Phase~3 routes the original image plus the JSON artefact to two MLLM reasoning agents, each producing a complete XAI report.
This is a two-stage information-fusion pipeline: Phase~1 performs \emph{decision-level fusion} implicitly available to Phase~3 (both CNN decisions are exposed side by side), and Phase~3 performs \emph{semantic-level fusion}, using pre-trained visual-linguistic knowledge to weigh, reconcile, or override the two decision-level signals.

\subsection{Phase 1: Perceptual Classification}

\subsubsection{EfficientNet-B3}

EfficientNet-B3 \cite{Tan2019EfficientNet} employs \emph{compound scaling} jointly increasing network depth, width, and input resolution to achieve an efficient accuracy-parameter trade-off.
Its design philosophy makes it particularly suited to fine-grained texture tasks where spatial resolution matters (e.g., spotting concentric ring patterns characteristic of early blight).

We adopt a \emph{partial-unfreeze} transfer learning strategy: the early feature-extraction blocks (\texttt{features[0--5]}) are frozen to preserve low-level ImageNet representations, while the top three blocks (\texttt{features[6--8]}), the adaptive pool, and the classification head are unfrozen.
This prevents catastrophic forgetting while allowing task-specific adaptation, and is applied identically on all three datasets.

\subsubsection{ConvNeXt-Tiny}

ConvNeXt-Tiny \cite{Liu2022ConvNeXt} modernises the classical ResNet paradigm by incorporating Vision Transformer design principles, large-kernel depthwise convolutions (7$\times$7), inverted bottleneck structure, LayerNorm, and GELU activations, while retaining the translation invariance and locality inductive biases of standard convolutions.
We adopt a \emph{full fine-tune} strategy with stochastic depth \cite{Cubuk2020RandAugment} (drop rate 0.1) and RandAugment \cite{Cubuk2020RandAugment} augmentation, which we found critical for generalisation on all three datasets, particularly PlantDoc's small training set.

The contrasting design philosophies (compound scaling vs. modernised convolution) are deliberate: architecturally distinct experts produce meaningfully different confidence distributions, creating genuine conflicts that motivate MLLM arbitration on PlantDoc, even though the same experts converge to near-unanimous agreement on the visually narrower Cornell datasets (Section~\ref{sec:Results}).
Table~\ref{tab:cnn_hparams} summarises the hyperparameters for both models, which are identical across all three datasets except for the number of output classes.

\begin{table*}[htb]
\centering
\caption{Fine-tuning hyperparameters for the two CNN perceptual experts, applied uniformly to PlantDoc and both Cornell datasets.}
\label{tab:cnn_hparams}
\small
\setlength{\tabcolsep}{5pt}
\renewcommand{\arraystretch}{1.15}

% Adjust the three p{...} values below to change the column widths.
\begin{tabular}{
    >{\raggedright\arraybackslash}p{0.22\textwidth}
    >{\raggedright\arraybackslash}p{0.34\textwidth}
    >{\raggedright\arraybackslash}p{0.34\textwidth}
}
\hline
\textbf{Parameter} &
\textbf{EfficientNet-B3} &
\textbf{ConvNeXt-Tiny} \\
\hline
Input resolution
& $300 \times 300$
& $224 \times 224$ \\

Pre-trained weights
& ImageNet-1k
& ImageNet-1k \\

Fine-tuning strategy
& Partial fine-tuning (top 30\%)
& Full-network fine-tuning \\

Optimizer
& AdamW
& AdamW \\

Learning rate
& $1 \times 10^{-4}$
& $4 \times 10^{-4}$ \\

Learning-rate schedule
& Cosine annealing
& Cosine annealing with 5-epoch warm-up \\

Minimum learning rate
& $1 \times 10^{-6}$
& $1 \times 10^{-6}$ \\

Weight decay
& $1 \times 10^{-2}$
& $5 \times 10^{-2}$ \\

Total batch size
& 256
& 256 \\

Computing hardware
& $4 \times$ NVIDIA H100 GPUs using DDP
& $4 \times$ NVIDIA H100 GPUs using DDP \\

Training epochs
& 50
& 50 \\

Loss function
& Focal Loss ($\gamma=2$)
& Focal Loss ($\gamma=2$) \\

Data augmentation
& Horizontal flipping, rotation, color jitter, and Gaussian blur
& Horizontal flipping, rotation, color jitter, Gaussian blur, and RandAugment \\
\hline
\end{tabular}
\end{table*}

\subsubsection{Training Infrastructure}

All CNN training runs (six in total: two architectures $\times$ three datasets) use PyTorch DistributedDataParallel (DDP) \cite{Li2020PyTorchDDP} across four NVIDIA H100 NVL GPUs (CUDA~12.8, PyTorch~2.x+cu124).
Focal Loss \cite{Lin2017FocalLoss} with inverse-frequency class weights is used to mitigate class imbalance, which is severe for PlantDoc (up to 6.6$\times$) and comparatively mild for the Cornell datasets (up to 2.9$\times$ between Late Blight and Early Blight in Stage~4).
Best checkpoints are selected based on validation Macro~F1, computed at the end of each epoch.

\subsection{Phase 2: Contextual Ingestion and Alignment}
\label{sec:contextual_ingestion}

As illustrated in Fig.~\ref{fig:framework} and Algorithm 1, Phase~2 serves as the contextual alignment layer between the CNN-based perceptual experts in Phase~1 and the MLLM-based semantic reasoning agents in Phase~3. Its purpose is to transform heterogeneous model outputs, image identifiers, and decision-relevant agricultural information into a standardized, machine-readable evidence representation. For every test image, the pipeline constructs a structured JSON artefact that records: (a) the image identifier or file path and its ground-truth class, where the ground truth is retained exclusively by the evaluation pipeline and is never provided to the MLLM; (b) the top-3 class predictions and corresponding confidence scores generated independently by EfficientNet-B3 and ConvNeXt-Tiny; and (c) a structured \emph{business context} object containing the contextual variables required for downstream risk interpretation and management-oriented reporting. This intermediate representation preserves the identity of each perceptual expert rather than prematurely averaging their probability distributions, thereby enabling the MLLM to identify agreement, quantify disagreement, compare confidence levels, and provide an explicit justification for accepting or overriding a CNN prediction.

The business-context schema is adapted to the structure of each dataset while maintaining a consistent JSON organization. For PlantDoc, the contextual object is generated from a predefined crop-to-value lookup table that maps the identified plant species to an estimated crop value in euros, a risk-aversion level, and the prevailing or assumed weather condition. For the Cornell Stage~2 and Stage~4 datasets, the context is disease-specific rather than species-specific because all images represent tomato plants. As described in Section~\ref{sec:dataset_cornell}, the Cornell schema therefore includes the candidate disease, its causal pathogen, the estimated crop value, the assigned risk-aversion level, and treatment urgency. These variables allow the subsequent MLLM layer to interpret the perceptual evidence in relation to the potential biological and operational consequences of the diagnosis rather than treating every predicted class as having equivalent management importance.

Algorithm 1 presents a representative JSON evidence artefact generated for a PlantDoc test image. For visual compactness, the displayed example shows the two highest-ranked predictions from each CNN, whereas the implemented pipeline retains the top-3 predictions and confidence scores. In this example, EfficientNet-B3 assigns its highest confidence to \texttt{Tomato\_Early\_blight\_leaf}, while ConvNeXt-Tiny assigns its highest confidence to \texttt{Tomato\_leaf}. The artefact preserves this disagreement and supplements it with the associated business context, allowing the MLLM to arbitrate between the competing perceptual signals using both visual and structured evidence.

The explicit grounding of MLLM reasoning in precomputed CNN evidence provides two principal advantages. First, it reduces the likelihood of unsupported diagnostic generation by directing the MLLM toward a finite set of candidate classes identified by the perceptual experts. Although the MLLM may override both CNNs when its visual interpretation strongly supports another diagnosis, such behavior becomes explicitly detectable and can be analyzed as a separate override category. Second, the artefact establishes a transparent and auditable evidence trail linking the final diagnosis to the predictions and confidence scores of both CNNs. Each MLLM decision can therefore be traced to expert agreement, confidence-weighted selection, semantic conflict resolution, or an explicit independent override.
\begin{tcblisting}{
    enhanced,
    breakable,
    colback=green!10,
    colframe=algorithmblue,
    coltext=black,
    coltitle=white,
    toprule=2.5pt,
    bottomrule=1.0pt,
    leftrule=1.0pt,
    rightrule=1.0pt,
    arc=2mm,
    left=2mm,
    right=2mm,
    top=2mm,
    bottom=1mm,
    before skip=10pt,
    after skip=10pt,
    title=\textbf{Algorithm 1: Structured JSON Evidence Artefact},
    fonttitle=\bfseries,
    attach boxed title to top left={
        xshift=4mm,
        yshift=-\tcboxedtitleheight/2
    },
    boxed title style={
        colback=algorithmblue,
        colframe=algorithmblue,
        coltext=white,
        toprule=2.5pt,
        bottomrule=1.0pt,
        leftrule=1.0pt,
        rightrule=1.0pt,
        arc=1mm,
        left=2mm,
        right=2mm
    },
    label={alg:json_evidence},
    listing only,
    listing options={
        language=json,
        basicstyle=\ttfamily\small\color{black},
        breaklines=true,
        showstringspaces=false,
        columns=fullflexible,
        keepspaces=true
    }
}
{
  "image_id": "test_Tomato_Early_blight_leaf_3.jpg",
  "vision_experts": {
    "EfficientNet-B3": [
      {
        "class": "Tomato_Early_blight_leaf",
        "confidence": 0.72
      },
      {
        "class": "Tomato_leaf",
        "confidence": 0.15
      }
    ],
    "ConvNeXt-Tiny": [
      {
        "class": "Tomato_leaf",
        "confidence": 0.58
      },
      {
        "class": "Tomato_Early_blight_leaf",
        "confidence": 0.36
      }
    ]
  },
  "business_context": {
    "crop_value": 50000,
    "current_weather": "variable",
    "risk_aversion": "high"
  }
}
\end{tcblisting}

This contextual alignment mechanism also separates perceptual inference from business-oriented interpretation. The CNN experts remain responsible for extracting discriminative visual evidence, whereas the MLLM evaluates their predictions in conjunction with the original image and structured agricultural context. The resulting modularity allows perceptual models, contextual records, or MLLM agents to be updated independently without redesigning the entire framework. More importantly, the JSON artefact provides a reproducible interface through which the behavior of different MLLMs can be compared under identical perceptual and contextual inputs.

\subsubsection{Prompt Design}

Careful prompt engineering is essential for reliable structured output from instruction-following MLLMs.
We use a two-part prompt structure, held constant across all three datasets apart from the class-name vocabulary and business-context fields.

\textbf{System prompt (Gemma~4~E4B, PlantDoc variant):}
\begin{quote}
\small\textit{You are an expert agricultural AI assistant specialising in plant disease diagnosis and business intelligence for crop management. You will receive a leaf image alongside predictions from two independent computer-vision models. Your role is to synthesise these predictions, resolve any disagreement using visual evidence, and deliver a precise, actionable business recommendation.}
\end{quote}

\textbf{System prompt (Qwen3.5~4B, condensed):}
\begin{quote}
\small\textit{You are a plant disease diagnosis assistant. You must respond with ONLY a JSON object. No thinking. No explanation. No markdown. Just the JSON.}
\end{quote}

For the Cornell datasets, the equivalent system prompts explicitly enumerate the three possible diagnoses (\texttt{Early\_Blight}, \texttt{Late\_Blight}, \texttt{Septoria\_Leaf\_Spot}) and note that all three are tomato foliar diseases, focusing the MLLM's semantic reasoning on the narrower, disease-specific vocabulary appropriate to that dataset.

The user prompt, identical in structure across datasets, presents: (i)~the leaf image; (ii)~the two CNN prediction blocks with class names and confidence scores; (iii)~the business context; and (iv)~a strict instruction to output only a JSON object with exactly seven fields.
For Qwen3.5, the condensed system prompt was found empirically to be necessary on every dataset: the full system prompt caused the model to enter its chain-of-thought reasoning mode, exhausting the 2,048-token generation budget with verbose analysis before reaching the JSON output.

\subsubsection{Output Validation and Error Recovery}

MLLM responses are validated with a two-pass parser: (1)~direct regex-based JSON extraction, and (2)~re-attempt after stripping Markdown code fence delimiters (\texttt{```json} / \texttt{```}).
Responses that fail both passes are recorded as parse errors; their images are excluded from accuracy computation but are counted against coverage.
In production deployment, we recommend implementing a retry loop with a simplified prompt as a third fallback, which would eliminate the small residual parse failures observed on each dataset (0.8--3.2\%, depending on model and dataset).

\subsection{Phase 3: Multimodal Cognitive Reasoning}

\subsubsection{MLLM Selection}

We select two small, open-weight MLLMs from distinct model families to demonstrate framework generalisability across datasets:

\textbf{Google Gemma~4~E4B} \cite{Gemma4_2025} employs Per-Layer Embeddings (PLE) for parameter efficiency, achieving 4.5B effective parameters despite 8B stored weights.
It supports text, image, and audio modalities with a 128K-token context window and a configurable thinking/reasoning mode.

\textbf{Alibaba Qwen3.5~4B} \cite{Qwen3_2025} is a dense, unified vision-language model with a 4B-parameter transformer backbone and a built-in visual encoder.
It supports text and image inputs with strong multimodal reasoning capabilities.
During inference, \texttt{enable\_thinking=False} is passed to \texttt{apply\_chat\_template} to suppress its chain-of-thought reasoning mode and force direct structured output.

Both models are loaded via the HuggingFace \texttt{transformers} library (\texttt{AutoModelForImageTextToText}) with \texttt{device\_map="auto"} on a single H100 GPU, for all three datasets.

\subsubsection{Prompt Engineering}

Each MLLM receives a \emph{system prompt} establishing its role as an agricultural AI expert, followed by a structured \emph{user prompt} comprising:
(i)~the leaf image;
(ii)~the formatted CNN predictions with confidence scores;
(iii)~the business context; and
(iv)~an explicit instruction to respond with only a valid JSON object containing seven mandatory fields: \texttt{final\_diagnosis}, \texttt{confidence}, \texttt{arbitration\_reasoning}, \texttt{visual\_symptoms\_observed}, \texttt{risk\_level}, \texttt{business\_recommendation}, and \texttt{treatment\_window\_hours}.
This seven-field schema is held fixed across all three datasets, enabling direct cross-dataset comparison of MLLM behaviour (Section~\ref{sec:Results}).

%% ─────────────────────────────────────────────────────────────────────────────
\section{Experimental Setup}
\label{sec:Setup}

All experiments, across all three datasets, are conducted on a compute server equipped with four NVIDIA H100 NVL GPUs (80~GB VRAM each), CUDA~12.8, and PyTorch~2.x compiled for CUDA~12.4 (cu124).
CNN fine-tuning runs for 50 epochs with checkpoint selection based on validation Macro~F1; wall-clock training time is approximately 2~hours per model on PlantDoc and 2--4~hours per model on the larger Cornell datasets.
MLLM inference is performed on a single H100 GPU with \texttt{device\_map="auto"} and \texttt{dtype="auto"} (BF16); each test image requires a single forward pass per MLLM.

\textbf{Hyperparameter selection.}
All CNN hyperparameters listed in Table~\ref{tab:cnn_hparams} were set based on established conventions for the respective architectures on small-data classification tasks and applied without dataset-specific tuning, in order to isolate the effect of dataset realism (rather than per-dataset hyperparameter optimisation) on the reported results.
The partial-unfreeze depth for EfficientNet-B3 (top 30\%, corresponding to \texttt{features[6--8]}) was chosen to preserve low-level texture detectors learned on ImageNet while adapting higher-level semantic representations to the disease classification task.
The full fine-tune strategy for ConvNeXt-Tiny is consistent with the original ConvNeXt paper's recommendation for transfer learning on small datasets, where stochastic depth and AdamW weight decay provide sufficient regularisation to prevent overfitting.

\textbf{MLLM generation parameters.}
Both MLLMs use \texttt{temperature=1.0}, \texttt{max\_new\_tokens=2048}, and \texttt{do\_sample=True}, held constant across all three datasets.
Temperature scaling parameters (\texttt{top\_p}, \texttt{top\_k}) are set to their respective model defaults to avoid interfering with the models' calibrated sampling distributions.

\textbf{Evaluation metrics.}
We report Top-1 accuracy, Macro~F1 (unweighted mean over all target classes), and Weighted~F1 (weighted by class support), computed independently for each dataset.
Macro~F1 is the primary metric because it is invariant to class imbalance and therefore reflects performance on minority disease classes, which are of greatest agronomic significance.

\textbf{MLLM output validation.}
MLLM responses are parsed with the two-pass JSON extractor described in Section~\ref{sec:Methodology}.
Responses that fail both passes are recorded as parse errors and excluded from accuracy computation (coverage is reported separately for each dataset and model).

\textbf{Cross-dataset comparability.}
Because the three datasets differ in class count (27 vs.\ 3), test-set size (252 vs.\ 403 vs.\ 715), and label taxonomy, absolute accuracy values are not directly comparable across datasets; instead, our cross-dataset synthesis (Section~\ref{sec:Results}) focuses on \emph{relative} quantities that are meaningful regardless of class count---the accuracy gain of MLLM arbitration over the best CNN, the vision-expert agreement rate, and the calibration of MLLM risk assignment against known class prevalence.

%% ─────────────────────────────────────────────────────────────────────────────
\section{Results}
\label{sec:Results}

Results are reported per dataset (Sections~\ref{sec:results_plantdoc}--\ref{sec:results_stage4}) followed by a cross-dataset synthesis (Section~\ref{sec:results_synthesis}) that constitutes the primary evidence for the paper's central claims regarding conflict-dependent MLLM utility and risk-personality calibration.

\subsection{PlantDoc Results}
\label{sec:results_plantdoc}

\subsubsection{Full Per-Class Classification Report}

Table~\ref{tab:fullperclass} reports per-class precision, recall, and F1 for ConvNeXt-Tiny and per-class accuracy for both MLLM agents (Gemma~4~E4B and Qwen3.5~4B) across all 27 classes.
The table highlights the heterogeneity of performance: six classes achieve F1~$\geq$~0.84 for ConvNeXt-Tiny (Strawberry, Grape Leaf, Grape Black Rot, Raspberry, Squash Powdery Mildew, Corn Rust), while six classes fall below F1~=~0.42 (Corn Gray Leaf Spot, Potato Early Blight, Potato Late Blight, Tomato Mold, Tomato Bacterial Spot, Tomato Mosaic Virus).

\begin{table*}[htb]
\centering
\caption{Full per-class results on PlantDoc. CNX = ConvNeXt-Tiny (Precision / Recall / F1); Gem = Gemma~4~E4B (top-1 accuracy); Qwn = Qwen3.5~4B (top-1 accuracy). $N$ = test support. Best accuracy per class in bold.}
\label{tab:fullperclass}
\scriptsize
\begin{tabular}{lcccc|cc}
\hline
\textbf{Class} & \textbf{P} & \textbf{R} & \textbf{F1} & $N$ & \textbf{Gem} & \textbf{Qwn} \\
\hline
Apple Scab Leaf          & 0.78 & 0.70 & 0.74 & 10 & \textbf{90\%} & 70\% \\
Apple Leaf               & \textbf{0.57} & \textbf{0.89} & \textbf{0.70} &  9 & \textbf{78\%} & 67\% \\
Apple Rust Leaf          & 1.00 & 0.60 & 0.75 & 10 & 60\% & \textbf{70\%} \\
Bell Pepper Leaf         & 0.80 & 0.50 & 0.62 &  8 & \textbf{62\%} & \textbf{62\%} \\
Bell Pepper Leaf Spot    & 0.45 & 0.56 & 0.50 &  9 & 56\% & \textbf{67\%} \\
Blueberry Leaf           & 0.88 & 0.64 & 0.74 & 11 & 45\% & \textbf{64\%} \\
Cherry Leaf              & 0.62 & 0.50 & 0.56 & 10 & 40\% & \textbf{50\%} \\
Corn Gray Leaf Spot      & 0.09 & 0.25 & 0.13 &  4 & \textbf{50\%} & \textbf{50\%} \\
Corn Leaf Blight         & 0.57 & 0.33 & 0.42 & 12 & 33\% & \textbf{33\%} \\
Corn Rust Leaf           & 1.00 & 0.80 & 0.89 & 10 & 80\% & 80\% \\
Peach Leaf               & 0.88 & 0.78 & 0.82 &  9 & \textbf{78\%} & \textbf{78\%} \\
Potato Early Blight      & 0.27 & 0.21 & 0.24 & 14 & 21\% & \textbf{21\%} \\
Potato Late Blight       & 0.31 & 0.50 & 0.38 &  8 & \textbf{50\%} & \textbf{50\%} \\
Raspberry Leaf           & 0.88 & 1.00 & 0.93 &  7 & \textbf{100\%} & 86\% \\
Soyabean Leaf            & 0.57 & 0.50 & 0.53 &  8 & 38\% & \textbf{50\%} \\
Squash Powdery Mildew    & 1.00 & 0.83 & 0.91 &  6 & \textbf{83\%} & 50\% \\
Strawberry Leaf          & 1.00 & 1.00 & 1.00 &  8 & \textbf{100\%} & 88\% \\
Tomato Early Blight      & 0.70 & 0.78 & 0.74 &  9 & \textbf{89\%} & \textbf{89\%} \\
Tomato Septoria Spot     & 0.56 & 0.83 & 0.67 & 12 & \textbf{92\%} & \textbf{92\%} \\
Tomato Leaf              & 0.75 & 0.38 & 0.50 &  8 & 38\% & \textbf{25\%} \\
Tomato Bacterial Spot    & 0.43 & 0.33 & 0.38 &  9 & 22\% & \textbf{44\%} \\
Tomato Late Blight       & 0.86 & 0.60 & 0.71 & 10 & 60\% & \textbf{60\%} \\
Tomato Mosaic Virus      & 0.50 & 0.30 & 0.38 & 10 & \textbf{50\%} & \textbf{60\%} \\
Tomato Yellow Virus      & 0.92 & 0.80 & 0.86 & 15 & 80\% & 80\% \\
Tomato Mold              & 0.24 & 0.67 & 0.35 &  6 & \textbf{67\%} & \textbf{67\%} \\
Grape Leaf               & 1.00 & 1.00 & 1.00 & 12 & \textbf{100\%} & \textbf{100\%} \\
Grape Black Rot          & 0.73 & 1.00 & 0.84 &  8 & \textbf{100\%} & \textbf{100\%} \\
\hline
\textbf{Macro avg}       & 0.68 & 0.64 & 0.64 & 252 & 68.5\% & 67.2\% \\
\hline
\end{tabular}
\end{table*}

The six hardest classes share a common characteristic: they exhibit high visual similarity to other classes within the same genus or infection type.
Potato Early Blight and Potato Late Blight, for example, both produce necrotic lesions on potato foliage; the distinguishing features (lesion shape, water-soaked border, sporulation pattern) require fine spatial resolution and semantic knowledge that the available training images (14 and 8 test samples respectively) do not sufficiently represent.
Tomato Mold (\textit{Botrytis cinerea}) and Tomato Bacterial Spot both produce small dark lesions, making textural discrimination especially difficult.

The Gemma column reveals that MLLM gains are concentrated in exactly these difficult classes: Tomato Septoria (+8.3\%), Tomato Early Blight (+11.1\%), Tomato Mosaic Virus (+20\%), Apple Scab (+20\%), and Corn Gray Leaf Spot (+25\%), while structurally distinctive, easier classes (Strawberry, Grape, Corn Rust) show no change since the CNN is already near-perfect.
Qwen3.5~4B shows a complementary pattern: it outperforms Gemma on Tomato Bacterial Spot (44\% vs.~22\%), Tomato Mosaic Virus (60\% vs.~50\%), and Blueberry Leaf (64\% vs.~45\%), but underperforms on Apple Scab (70\% vs.~90\%) and Squash Powdery Mildew (50\% vs.~83\%), reinforcing that different MLLM families encode different disease-specific strengths.

Table~\ref{tab:main_results} presents the primary quantitative comparison across all four models on the PlantDoc held-out test set.

\begin{table*}[htb]
\centering
\caption{Classification performance on the PlantDoc test set (252 images, 27 classes). Coverage: fraction of test images for which the model produced a parseable prediction. Best results in \textbf{bold}.}
\label{tab:main_results}
\begin{tabular}{lcccc}
\hline
\textbf{Model} & \textbf{Acc.} & \textbf{Macro F1} & \textbf{W. F1} & \textbf{Cov.} \\
\hline
EfficientNet-B3   & 59.1\% & 0.5857 & 0.5872 & 100\% \\
ConvNeXt-Tiny     & 63.9\% & 0.6393 & 0.6479 & 100\% \\
\hline
Gemma 4 E4B (MLLM)& \textbf{68.5\%} & \textbf{0.6843} & \textbf{0.6906} & 94.4\% \\
Qwen3.5 4B (MLLM) & 67.2\% & 0.6761 & 0.6788 & 96.8\% \\
\hline
\end{tabular}
\end{table*}

Both MLLM agents outperform both CNN baselines.
Gemma~4~E4B achieves the highest accuracy (68.5\%) and Macro~F1 (0.6843), representing a +9.4-point gain over EfficientNet-B3 and a +4.6-point gain over ConvNeXt-Tiny.
Qwen3.5~4B follows closely at 67.2\% (+8.1 and +3.3 points respectively).

\subsubsection{Contextualisation within the PlantDoc Literature}

Table~\ref{tab:literature_comparison} summarizes accuracies reported by recent studies on the same or comparable PlantDoc splits.

\begin{table*}[htb]
\centering
\caption{PlantDoc classification accuracy reported in recent literature. ``Full split'' indicates use of the official 27-class train/test split without class-balanced subsetting or external data.}
\label{tab:literature_comparison}
\scriptsize
\begin{tabular}{lccc}
\hline
\textbf{Study} & \textbf{Model(s)} & \textbf{Acc.} & \textbf{Full split?} \\
\hline
Singh et al. \cite{Singh_2020}        & VGG-16, ResNet-50          & 30--50\% & Partial \\
Taneja et al. \cite{Taneja2023Comparative} & Custom CNN, VGG-16     & 60--70\% & Partial \\
Shiyan et al. \cite{Shiyan2025Recognizing}  & MobileNetV3, Eff-B0, DenseNet-121 & 60--75\% & Yes \\
Fatma et al. \cite{Fatma2025YOLOv8}         & YOLOv8, EfficientNet-B0 & 65--75\% & Yes \\
Wojciuk et al. \cite{Wojciuk2024Improving}  & Fine-tuned CNNs (HPO)   & 70--80\% & No (balanced subset) \\
Hasan et al. \cite{Hasan2025ECA}            & YOLOv11n + ECA-NFNet    & 75--80\% & No (two-stage) \\
Chettri et al. \cite{Chettri2025ResNetSwin} & ResNet-50 + Swin-T      & 75--80\% & No (hybrid + weighted) \\
\hline
\textbf{This work} (EfficientNet-B3)        & EfficientNet-B3          & 59.1\%  & \textbf{Yes} \\
\textbf{This work} (ConvNeXt-Tiny)          & ConvNeXt-Tiny            & 63.9\%  & \textbf{Yes} \\
\textbf{This work} (Gemma 4 E4B)            & Gemma 4 E4B (zero-shot)  & \textbf{68.5\%}  & \textbf{Yes} \\
\textbf{This work} (Qwen3.5 4B)             & Qwen3.5 4B (zero-shot)   & \textbf{67.2\%}  & \textbf{Yes} \\
\hline
\end{tabular}
\end{table*}

Our CNN baselines achieve 59--64\% accuracy on the full 27-class PlantDoc test split, consistent with the 60--75\% range reported in the literature for single-CNN architectures on this challenging in-the-wild benchmark \cite{Shiyan2025Recognizing,Fatma2025YOLOv8}.
Studies reporting higher accuracies (75--80\%) typically employ strategies that make direct comparison difficult: class-balanced subsetting \cite{Wojciuk2024Improving}, two-stage detection-then-classification pipelines \cite{Hasan2025ECA}, or hybrid CNN+Transformer architectures with weighted sampling \cite{Chettri2025ResNetSwin}.
Xiang et al. \cite{Xiang2026Reliability} recently confirmed that the controlled-to-field reliability gap remains severe even with standard mitigation techniques, reinforcing that PlantDoc's full-split accuracy ceiling for single-CNN approaches is approximately 65--75\%.
The MLLM arbitration layer improves accuracy to 68.5\%, competitive with hybrid and two-stage approaches, while requiring no PlantDoc-specific MLLM training and no additional architectural complexity beyond the two CNN backbones.

\subsubsection{CNN Validation Performance}

On the held-out validation set (401 images), ConvNeXt-Tiny achieves a Macro~F1 of 0.7928, substantially higher than EfficientNet-B3's 0.6594.
This 13.3-point validation gap confirms that the two perceptual experts have genuinely different capability profiles---a prerequisite for meaningful conflict analysis.

\subsubsection{Vision Expert Agreement and Conflict Analysis}

\begin{table}[htb]
\centering
\caption{PlantDoc: accuracy conditioned on vision expert agreement. All four models are evaluated separately on the 147 consensus images and the 105 conflict images.}
\label{tab:agreement}
\begin{tabular}{lcc}
\hline
\textbf{Model} & \textbf{Agree (147)} & \textbf{Disagree (105)} \\
\hline
EfficientNet-B3    & 78.2\% & 32.4\% \\
ConvNeXt-Tiny      & 78.2\% & 43.8\% \\
Gemma 4 E4B (MLLM) & 76.9\% & \textbf{47.6\%} \\
Qwen3.5 4B (MLLM)  & 75.9\% & \textbf{51.4\%} \\
\hline
\end{tabular}
\end{table}

Out of 252 test images, the two CNN experts produce the same top-1 prediction on 147 (58.3\%) and disagree on 105 (41.7\%).
Table~\ref{tab:agreement} reveals a pivotal finding: on \emph{consensus} images, all four models perform similarly ($\approx$77--78\%), whereas on \emph{conflict} images, the two MLLMs achieve 47.6\% (Gemma) and 51.4\% (Qwen)---substantially outperforming EfficientNet-B3 (32.4\%) and ConvNeXt-Tiny (43.8\%).
Qwen3.5~4B's 7.6-point gain over the stronger CNN on conflict images is a headline result of this dataset: the MLLM reasoning layer adds the most value precisely where the visual evidence is most ambiguous.
As shown in Section~\ref{sec:results_synthesis}, this conflict rate (41.7\%) is an order of magnitude higher than on either Cornell dataset, and the conflict-dependent nature of this gain is a central cross-dataset finding of this paper.

\subsubsection{MLLM Override Behaviour}

\begin{table}[htb]
\centering
\caption{PlantDoc: MLLM override behaviour and accuracy per override type.}
\label{tab:override}
\begin{tabular}{lcccc}
\hline
\textbf{Override type} & \multicolumn{2}{c}{\textbf{Gemma 4 E4B}} & \multicolumn{2}{c}{\textbf{Qwen3.5 4B}} \\
 & Cases & Acc. & Cases & Acc. \\
\hline
Agrees with both CNNs   & 138 & 81.2\% & 138 & 79.7\% \\
Agrees w/ ConvNeXt only & 80  & 50.0\% & 83  & 50.6\% \\
Agrees w/ EfficientNet  & 12  & 66.7\% & 12  & 75.0\% \\
Overrides both CNNs     & 22  & 13.6\% & 17  & 17.6\% \\
\hline
\end{tabular}
\end{table}

Table~\ref{tab:override} decomposes MLLM decisions into four categories.
When the MLLM aligns with both CNN experts (138 cases), accuracy reaches 79.7--81.2\%, confirming that consensus among all three models is highly reliable.
Most critically, when the MLLM \emph{overrides both CNNs} with an independent prediction (22 cases for Gemma, 17 for Qwen), accuracy collapses to 13.6\%--17.6\%.
This finding establishes an important design principle, revisited for the Cornell datasets in Section~\ref{sec:results_synthesis}: \emph{MLLMs should arbitrate between CNN signals rather than generate independent visual classifications}.

\subsubsection{Arbitration Benefit, Rescue, and Damage Cases}

\begin{table*}[htb]
\centering
\caption{PlantDoc: per-image outcomes when comparing each MLLM against the best available CNN prediction.}
\label{tab:benefit}
\begin{tabular}{lcc}
\hline
\textbf{Outcome} & \textbf{Gemma} & \textbf{Qwen} \\
\hline
MLLM improves over best CNN  &  3 cases (1.2\%) &  3 cases (1.2\%) \\
MLLM hurts vs. best CNN      & 35 cases (13.9\%)& 32 cases (12.8\%)\\
Neutral (same outcome)       &214 cases (84.9\%)&215 cases (86.0\%)\\
\hline
Rescue rate (both CNNs wrong)& 5.3\% (3/57)    & 5.3\% (3/57) \\
Damage rate (both CNNs right)& 2.6\% (3/115)   & 2.6\% (3/115) \\
\hline
\end{tabular}
\end{table*}

Table~\ref{tab:benefit} shows that the primary mechanism behind MLLM accuracy gains is not heroic recovery of cases where both CNNs fail (the rescue rate is only 5.3\%), but rather asymmetric performance on \emph{partially-correct} scenarios, where one CNN is right and the MLLM aligns with the correct expert.
The neutral rate of 85\% confirms that the MLLM largely preserves CNN signal fidelity; its damage rate (13.9\%) is the principal limitation.

\subsubsection{Confidence-Calibrated Performance}

\begin{table}[htb]
\centering
\caption{PlantDoc: accuracy conditioned on the maximum CNN confidence score, a proxy for image difficulty.}
\label{tab:confidence}
\begin{tabular}{lccc}
\hline
\textbf{Confidence bucket} & \textbf{Images} & \textbf{ConvNeXt} & \textbf{Gemma} \\
\hline
High ($\geq$0.8)      & 110 & 81.8\% & 80.9\% \\
Medium (0.5--0.8)     &  94 & 57.4\% & 57.4\% \\
Low ($<$0.5)          &  48 & 35.4\% & \textbf{41.7\%} \\
\hline
\end{tabular}
\end{table}

Table~\ref{tab:confidence} shows Gemma's largest advantage over ConvNeXt-Tiny is concentrated on low-confidence images (+6.3~points), precisely where the CNN visual signal is weakest.

% \subsubsection{Business Risk Distribution}
\subsubsection{Agronomic Risk-Level Distribution for Crop Management}

\begin{table*}[htb]
\centering
\caption{PlantDoc: business risk level distribution assigned by the MLLM reasoning agents across 252 test images.}
\label{tab:risk}
\begin{tabular}{lcccc}
\hline
\textbf{Model} & \textbf{Critical} & \textbf{High} & \textbf{Medium} & \textbf{Low} \\
\hline
Gemma 4 E4B & 25 (9.9\%)  & 99 (39.3\%) & 87 (34.5\%) & 41 (16.3\%) \\
Qwen3.5 4B  & 92 (36.5\%) & 58 (23.0\%) & 51 (20.2\%) & 49 (19.4\%) \\
\hline
\end{tabular}
\end{table*}

Table~\ref{tab:risk} reveals a striking behavioural difference: Qwen3.5~4B assigns 3.7$\times$ more Critical-risk labels than Gemma~4~E4B.
This finding, and whether it replicates on the Cornell datasets, is examined systematically in Section~\ref{sec:results_synthesis}.

\subsection{Cornell Stage~2 Results (Real-World Field Data, 20~GB)}
\label{sec:results_stage2}

\subsubsection{Classification Performance}

Table~\ref{tab:stage2_main} presents the primary quantitative comparison on the Stage~2 held-out test set (403 images, 3 classes).

\begin{table*}[htb]
\centering
\caption{Cornell Stage~2: classification performance (403 test images, 3 classes). Best results in \textbf{bold}.}
\label{tab:stage2_main}
\small
\begin{tabular}{lcccc}
\hline
\textbf{Model} & \textbf{Acc.} & \textbf{Macro F1} & \textbf{W. F1} & \textbf{Cov.} \\
\hline
EfficientNet-B3   & 98.5\% & 0.9852 & 0.9852 & 100\% \\
ConvNeXt-Tiny     & \textbf{99.8\%} & \textbf{0.9974} & \textbf{0.9975} & 100\% \\
\hline
Gemma 4 E4B (MLLM)& 99.3\% & 0.9921 & 0.9925 & 100\% \\
Qwen3.5 4B (MLLM) & 97.7\% & 0.9763 & 0.9773 & 98.5\% \\
\hline
\end{tabular}
\end{table*}

Unlike PlantDoc, ConvNeXt-Tiny alone is the top performer on Stage~2 (99.8\%), narrowly exceeding Gemma (99.3\%); Qwen trails at 97.7\%.
Table~\ref{tab:stage2_perclass} confirms that all models are essentially saturated on Early Blight and Late Blight, with the small residual error concentrated on Septoria Leaf Spot.

\begin{table}[htb]
\centering
\caption{Cornell Stage~2: per-class precision/recall/F1 (ConvNeXt-Tiny) and Gemma accuracy. $N$ = test support.}
\label{tab:stage2_perclass}
\small
\begin{tabular}{lcccc}
\hline
\textbf{Class} & \textbf{P} & \textbf{R} & \textbf{F1} & $N$ \\
\hline
Early Blight        & 1.00 & 1.00 & 1.00 & 147 \\
Late Blight         & 0.99 & 1.00 & 1.00 & 141 \\
Septoria Leaf Spot  & 1.00 & 0.99 & 1.00 & 115 \\
\hline
\end{tabular}
\end{table}

\subsubsection{Vision Expert Agreement and MLLM Behaviour}

The two CNN experts agree on 396 of 403 images (98.3\%) and disagree on only 7 (1.7\%)---an order of magnitude lower conflict rate than PlantDoc's 41.7\%.On the 396 consensus images, all four models achieve 99.0--100\% accuracy. On the 7 conflict images, however, ConvNeXt-Tiny alone is the strongest (85.7\%), while EfficientNet-B3 and Qwen3.5~4B each achieve only 14.3\% and Gemma achieves 57.1\%---the small sample size (n=7) makes this subset highly sensitive to individual errors, and we report it as an important boundary condition: \emph{with a near-zero conflict rate, there are too few disagreement cases for MLLM arbitration to provide a statistically reliable benefit, and in this particular sample the stronger CNN alone outperforms both MLLMs}.

MLLM override analysis (Table~\ref{tab:stage2_override}) confirms the pattern seen on PlantDoc: when either MLLM agrees with both CNN experts, accuracy is 100\%, but when an MLLM overrides both CNNs with an independent prediction the outcome is consistently poor---Qwen does so on 8 images and achieves 0\% accuracy on them, while Gemma never overrides both CNNs on this dataset.

\begin{table*}[htb]
\centering
\caption{Cornell Stage~2: MLLM override behaviour and accuracy per override type.}
\label{tab:stage2_override}
\small
\begin{tabular}{lcccc}
\hline
\textbf{Override type} & \multicolumn{2}{c}{\textbf{Gemma}} & \multicolumn{2}{c}{\textbf{Qwen}} \\
 & Cases & Acc. & Cases & Acc. \\
\hline
Agrees with both CNNs   & 396 & 100.0\% & 387 & 100.0\% \\
Agrees w/ ConvNeXt only & 5   & 80.0\%  & 1   & 100.0\% \\
Agrees w/ EfficientNet  & 2   & 0.0\%   & 2   & 0.0\% \\
Overrides both CNNs     & 0   & N/A     & 8   & 0.0\% \\
\hline
\end{tabular}
\end{table*}

Arbitration benefit analysis shows Gemma improves over the best CNN on none of the images, hurts on 0.7\% (3/403), and is neutral on 99.3\%; Qwen likewise improves on none, hurts on 2.5\% (10/403), and is neutral on 97.5\%.
There are no cases where both CNNs are wrong (rescue rate undefined); of the 396 images where both CNNs are correct, Gemma introduces no errors (0\% damage rate) while Qwen introduces 4 errors (1.0\% damage rate).

\subsubsection{Business Risk Distribution}

\begin{table*}[htb]
\centering
\caption{Cornell Stage~2: business risk level distribution across 403 test images.}
\label{tab:stage2_risk}
\small
\begin{tabular}{lccc}
\hline
\textbf{Model} & \textbf{Critical} & \textbf{High} & \textbf{Medium/Low/Unk.} \\
\hline
Gemma 4 E4B & 139 (34.5\%) & 78 (19.4\%) & 186 (46.2\%) / 0 / 0 \\
Qwen3.5 4B  & 127 (31.5\%) & 25 (6.2\%)  & 230 (57.1\%) / 16 (4.0\%) / 5 (1.2\%) \\
\hline
\end{tabular}
\end{table*}

Notably, on Stage~2 the Qwen-Gemma Critical-rate gap essentially disappears (34.5\% vs.\ 31.5\%), in sharp contrast to PlantDoc's 3.7$\times$ divergence.
As Late Blight---the sole class flagged \emph{high} risk-aversion in the business-context lookup table (Section~\ref{sec:dataset_cornell})---constitutes 141/403 = 35.0\% of this test set, both models' Critical rates are almost perfectly aligned with true urgent-disease prevalence.
This observation motivates the systematic risk-calibration analysis in Section~\ref{sec:results_synthesis}.

\subsection{Cornell Stage~4 Results (Real-World Field Data, 40~GB)}
\label{sec:results_stage4}

\subsubsection{Classification Performance}

Table~\ref{tab:stage4_main} reports results on the Stage~4 held-out test set (715 images, 3 classes)---a near-doubling of the Stage~2 dataset's volume.

\begin{table*}[htb]
\centering
\caption{Cornell Stage~4: classification performance (715 test images, 3 classes). Best results in \textbf{bold}.}
\label{tab:stage4_main}
\small
\begin{tabular}{lcccc}
\hline
\textbf{Model} & \textbf{Acc.} & \textbf{Macro F1} & \textbf{W. F1} & \textbf{Cov.} \\
\hline
EfficientNet-B3   & 96.8\% & 0.9725 & 0.9679 & 100\% \\
ConvNeXt-Tiny     & 98.3\% & 0.9856 & 0.9833 & 100\% \\
\hline
Gemma 4 E4B (MLLM)& \textbf{98.9\%} & \textbf{0.9900} & \textbf{0.9888} & 100\% \\
Qwen3.5 4B (MLLM) & 91.8\% & 0.9108 & 0.9200 & 98.9\% \\
\hline
\end{tabular}
\end{table*}

Gemma performs the best again on Stage~4 (98.9\%, matching the PlantDoc pattern), while Qwen's accuracy drops noticeably (91.8\%) relative to its Stage~2 performance (97.7\%) driven, as shown below, by a substantially higher override rate on the larger dataset. Table~\ref{tab:stage4_perclass} shows per-class performance remains strong across all three diseases, with Early Blight showing the largest residual error (96--98\% F1) among the three.

\begin{table}[htb]
\centering
\caption{Cornell Stage~4: per-class precision/recall/F1 (ConvNeXt-Tiny). $N$ = test support.}
\label{tab:stage4_perclass}
\small
\begin{tabular}{lcccc}
\hline
\textbf{Class} & \textbf{P} & \textbf{R} & \textbf{F1} & $N$ \\
\hline
Early Blight        & 0.99 & 0.97 & 0.98 & 329 \\
Late Blight         & 1.00 & 1.00 & 1.00 & 146 \\
Septoria Leaf Spot  & 0.96 & 0.99 & 0.98 & 240 \\
\hline
\end{tabular}
\end{table}

\subsubsection{Vision Expert Agreement and MLLM Behaviour}

The two CNN experts agree on 686 of 715 images (95.9\%) and disagree on 29 (4.1\%)---still far below PlantDoc's 41.7\%, but 2.4$\times$ higher than Stage~2's 1.7\%, consistent with Stage~4's larger, more visually diverse image pool.
On the 686 consensus images, EfficientNet-B3, ConvNeXt-Tiny, and Gemma each achieve 99.6\%, while Qwen trails at 93.2\%.
On the 29 conflict images, Gemma is the clear best performer at 82.8\%, a +13.8-point improvement over the stronger CNN (ConvNeXt-Tiny, 69.0\%); EfficientNet-B3 achieves only 31.0\% and Qwen 53.6\%.
This is the strongest positive conflict-arbitration result observed on any dataset in this study, and---unlike Stage~2's n=7 conflict subset---is based on a moderately-sized sample (n=29) that provides reasonable statistical support.

\begin{table}[htb]
\centering
\caption{Cornell Stage~4: MLLM override behaviour and accuracy per override type.}
\label{tab:stage4_override}
\small
\begin{tabular}{lcccc}
\hline
\textbf{Override type} & \multicolumn{2}{c}{\textbf{Gemma}} & \multicolumn{2}{c}{\textbf{Qwen}} \\
 & Cases & Acc. & Cases & Acc. \\
\hline
Agrees with both CNNs   & 686 & 99.6\% & 634 & 99.8\% \\
Agrees w/ ConvNeXt only & 19  & 89.5\% & 13  & 76.9\% \\
Agrees w/ EfficientNet  & 9   & 77.8\% & 7   & 71.4\% \\
Overrides both CNNs     & 1   & 0.0\%  & 54  & 1.9\% \\
\hline
\end{tabular}
\end{table}

Table~\ref{tab:stage4_override} reveals a crucial cross-dataset validation of the PlantDoc override-collapse finding: Qwen overrides both CNNs on 54 of 715 images (7.6\% of the test set, a much higher rate than on PlantDoc or Stage~2), and accuracy on exactly these cases collapses to 1.9\%.
This directly explains Qwen's lower aggregate accuracy on Stage~4 relative to Stage~2: a larger, more diverse image pool appears to trigger more frequent unconstrained overrides, each of which is overwhelmingly likely to be wrong.
Gemma, by contrast, overrides both CNNs on only 1 image, essentially eliminating this failure mode.

Arbitration benefit analysis shows Gemma improves over the best CNN on none of the images, hurts on 0.7\% (5/715), and is neutral on 99.3\%; Qwen improves on 0.1\% (1/715), hurts on 8.1\% (57/715), and is neutral on 91.8\%.
Of 3 images where both CNNs are wrong, Qwen rescues 1 (33.3\%) and Gemma rescues none; of 683 images where both CNNs are correct, Gemma introduces no errors (0\% damage) while Qwen introduces 44 (6.4\% damage)---by far the highest damage rate observed for any model on any dataset in this study, again attributable to Qwen's elevated override rate.

\subsubsection{Business Risk Distribution}

\begin{table*}[htb]
\centering
\caption{Cornell Stage~4: business risk level distribution across 715 test images.}
\label{tab:stage4_risk}
\small
\begin{tabular}{lccc}
\hline
\textbf{Model} & \textbf{Critical} & \textbf{High} & \textbf{Medium/Low/Unk.} \\
\hline
Gemma 4 E4B & 147 (20.6\%) & 229 (32.0\%) & 339 (47.4\%) / 0 / 0 \\
Qwen3.5 4B  & 249 (34.8\%) & 147 (20.6\%) & 306 (42.8\%) / 5 (0.7\%) / 8 (1.1\%) \\
\hline
\end{tabular}
\end{table*}

On Stage~4, the Qwen-alarmist pattern re-emerges strongly: Qwen's Critical rate (34.8\%) is 1.7$\times$ Gemma's (20.6\%).
Notably, Gemma's Critical rate (20.6\%) again closely tracks the true Late Blight prevalence in this test set (146/715 = 20.4\%), while Qwen substantially over-flags relative to this same ground truth---a pattern examined quantitatively next.

\subsection{Cross-Dataset Synthesis}
\label{sec:results_synthesis}

Having presented per-dataset results, we now synthesise the three experiments to test the paper's central claims: that MLLM arbitration utility scales with the vision-expert conflict rate, and that Gemma and Qwen exhibit consistent, quantifiable, and distinct risk-assessment behaviors.

\subsubsection{Accuracy Across Datasets}

\begin{figure*}[htb]
\centering
\includegraphics[width=\linewidth]{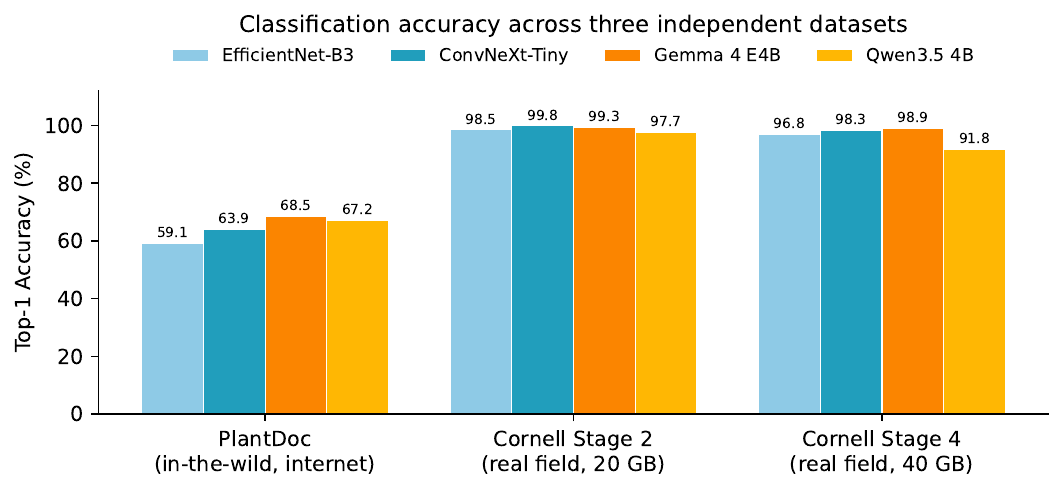}
\caption{Top-1 accuracy of all four models across the three evaluation datasets. All models improve substantially from PlantDoc's 27-class, internet-curated benchmark to the 3-class, real-world Cornell field datasets, with the CNN experts and Gemma remaining stable across both Cornell campaigns while Qwen shows greater variability.}
\label{fig:cross_accuracy}
\end{figure*}

Fig.~\ref{fig:cross_accuracy} plots Top-1 accuracy for all four models across the three datasets.
All models improve dramatically from PlantDoc (59.1--68.5\%) to the Cornell datasets (91.8--99.8\%), consistent with the taxonomic - difficulty and temporal-correlation factors discussed in Section~\ref{sec:dataset_cornell}.
Within the Cornell datasets, accuracy is remarkably stable for the CNN experts and Gemma (96.8--99.8\%), while Qwen shows more variability (97.7\% on Stage~2 vs.\ 91.8\% on Stage~4), consistent with its higher override rate on the larger, more diverse Stage~4 test set.

\subsubsection{Conflict-Dependent MLLM Utility}

\begin{figure*}[htb]
\centering
\includegraphics[width=\linewidth]{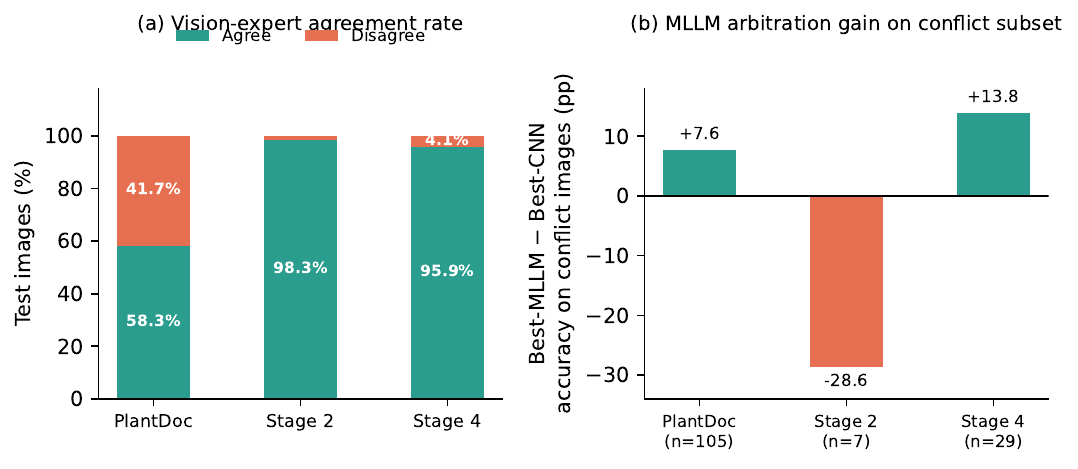}
\caption{(a) Vision-expert agreement/disagreement rate across the three datasets. (b) Best-MLLM minus best-CNN accuracy on the conflict-only image subset for each dataset, with sample size $n$ annotated. MLLM arbitration provides a substantial, positive gain when the conflict subset is reasonably sized (PlantDoc, $n=105$; Stage~4, $n=29$), but the tiny Stage~2 conflict subset ($n=7$) is dominated by sampling noise and should not be interpreted as evidence against the framework.}
\label{fig:conflict_utility}
\end{figure*}

Table~\ref{tab:synthesis_conflict} and Fig.~\ref{fig:conflict_utility} consolidate the vision-expert agreement rate and MLLM arbitration gain on conflict images across all three datasets---the paper's central cross-dataset finding.

\begin{table*}[htb]
\centering
\caption{Cross-dataset synthesis: vision-expert agreement/disagreement rate and best-MLLM vs.\ best-CNN accuracy on the conflict-only subset.}
\label{tab:synthesis_conflict}
\begin{tabular}{lccccc}
\hline
\textbf{Dataset} & \textbf{Test $N$} & \textbf{Agree} & \textbf{Disagree ($n$)} & \textbf{Best CNN (conflict)} & \textbf{Best MLLM (conflict)} \\
\hline
PlantDoc        & 252 & 58.3\% & 41.7\% (105) & 43.8\% (ConvNeXt) & \textbf{51.4\%} (Qwen), $+$7.6~pp \\
Cornell Stage~2 & 403 & 98.3\% &  1.7\% (7)   & \textbf{85.7\%} (ConvNeXt) & 57.1\% (Gemma), $-$28.6~pp \\
Cornell Stage~4 & 715 & 95.9\% &  4.1\% (29)  & 69.0\% (ConvNeXt) & \textbf{82.8\%} (Gemma), $+$13.8~pp \\
\hline
\end{tabular}
\end{table*}

The results in Table~\ref{tab:synthesis_conflict} support a nuanced conclusion.
Where the conflict subset is reasonably sized (PlantDoc, $n=105$; Stage~4, $n=29$), MLLM arbitration provides a clear positive gain over the stronger CNN (+7.6 and +13.8 points respectively), and this gain scales with the underlying conflict rate: Stage~4's 4.1\% conflict rate is an order of magnitude below PlantDoc's 41.7\%, yet still produces a larger \emph{relative} improvement (+13.8 vs.\ +7.6 points) on its (smaller) conflict subset, suggesting that Gemma's arbitration quality, not only the opportunity for arbitration, has improved on the more realistic Cornell imagery.

Where the conflict subset is very small (Stage~2, $n=7$), the observed $-$28.6-point ``loss'' is likely a statistical artifact of an extremely small sample rather than evidence that MLLM arbitration is harmful; a 7-image subset cannot support a reliable conclusion in either direction.
The overarching finding across all three datasets is therefore: \emph{MLLM arbitration utility is a function of the vision-expert conflict rate}---it is large and reliable when conflicts are frequent enough to be statistically meaningful (PlantDoc), remains positive and meaningful at moderate conflict rates even on unfamiliar, real-world imagery (Stage~4), and is simply not evaluable when conflicts are too rare to sample (Stage~2).

\subsubsection{Override Collapse: A Generalisable Design Principle}

The unconstrained-override collapse first identified on PlantDoc (13.6--17.6\% accuracy) replicates on Stage~4 (0--1.9\% accuracy over 1 and 54 override cases respectively) and is directionally consistent (though based on very few cases) on Stage~2 (0\% accuracy over 0 and 8 cases).
Across all three datasets and both MLLMs, overriding both CNN experts simultaneously never exceeds 17.6\% accuracy, and on the two Cornell datasets it is at or near 0\%.
This is, to our knowledge, the first replication of this design-critical finding across independent datasets of different scale, class count, and acquisition modality, and it substantially strengthens the general design principle first proposed in the context of PlantDoc alone: \emph{MLLMs deployed as arbitrators in a hierarchical fusion pipeline should be constrained to select among CNN-proposed candidates rather than permitted to generate independent diagnoses}.

\subsubsection{Risk-Personality Calibration Across Datasets}

\begin{figure*}[htb]
\centering
\includegraphics[width=\linewidth]{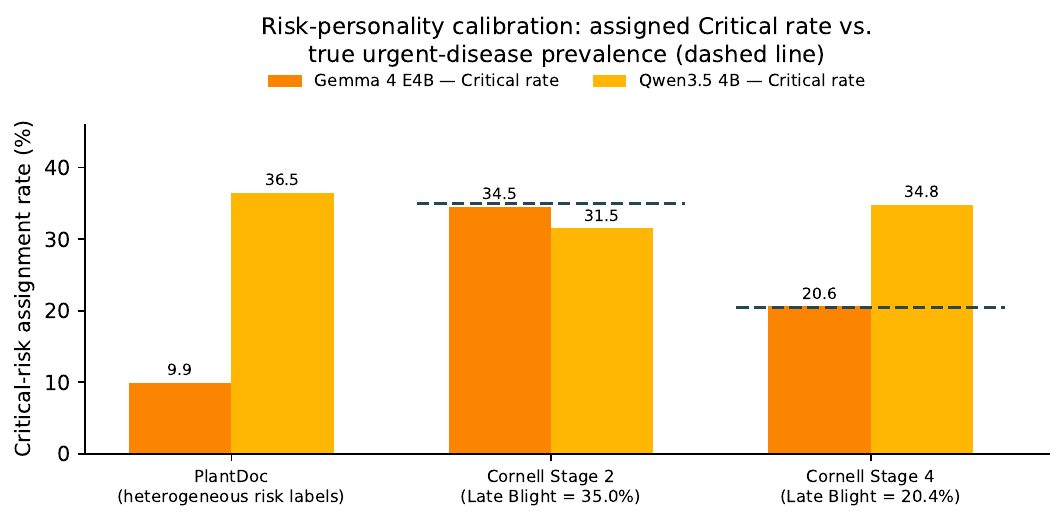}
\caption{Critical-risk assignment rate for Gemma~4~E4B and Qwen3.5~4B across the three datasets. Dashed lines indicate the true prevalence of the sole ``high risk-aversion'' disease class (Late Blight) in the two Cornell test sets, providing an objective calibration target. Gemma's Critical rate tracks true prevalence almost exactly (0.14--0.5-point error); Qwen consistently over-flags (3.5--14.4-point error).}
\label{fig:risk_calibration}
\end{figure*}

Fig.~\ref{fig:risk_calibration} presents the paper's second central cross-dataset finding: a quantitative \emph{risk-prevalence calibration} analysis comparing each MLLM's assigned Critical-risk rate against the true prevalence of Late Blight the only disease class flagged \emph{high} risk-aversion in the business-context lookup table (Section~\ref{sec:dataset_cornell}) on the two Cornell datasets, where this ground-truth comparison is well defined.

Table~\ref{tab:calibration_error} reports the resulting \emph{Critical-Risk Calibration Error} (CRCE)---the absolute difference between a model's assigned Critical-risk rate and the true prevalence of the urgent disease class---for both MLLMs on both Cornell datasets.

\begin{table*}[htb]
\centering
\caption{Critical-Risk Calibration Error (CRCE): absolute difference between assigned Critical-risk rate and true Late Blight prevalence, on the two Cornell datasets where ground truth is well defined.}
\label{tab:calibration_error}
\small
\begin{tabular}{lccc}
\hline
\textbf{Dataset} & \textbf{True prevalence} & \textbf{Gemma CRCE} & \textbf{Qwen CRCE} \\
\hline
Cornell Stage~2 & 35.0\% & \textbf{0.5~pp} & 3.5~pp \\
Cornell Stage~4 & 20.4\% & \textbf{0.14~pp} & 14.4~pp \\
\hline
\end{tabular}
\end{table*}

The result is fully consistent across both Cornell datasets: Gemma's Critical-risk rate is calibrated to within \emph{half a percentage point} of true urgent-disease prevalence on Stage~2 and within \emph{0.14 points} on Stage~4, while Qwen's error is 7--100$\times$ larger (3.5 and 14.4 points respectively).
This is not attributable to differences in classification accuracy alone Qwen's raw accuracy is close to Gemma's on Stage~2 (97.7\% vs.\ 99.3\%) but reflects a genuine, systematic difference in how each model translates diagnostic evidence into a risk category.
On PlantDoc, where 27 heterogeneous classes map to a more complex, species-level risk-aversion schema without a single dominant ``urgent'' class, both models' Critical rates diverge much further from any single reference value (Gemma 9.9\%, Qwen 36.5\%), and the 3.7$\times$ gap first reported for PlantDoc is best interpreted as the same underlying personality difference operating in a less-constrained, higher-dimensional risk-labelling regime.
Taken together, these results demonstrate that Qwen's ``alarmist'' tendency and Gemma's better calibration are not PlantDoc-specific artefacts but a consistent, quantifiable, model-level property that replicates across three independent datasets and two different risk-labelling regimes a finding with direct, actionable implications for which MLLM to select as the risk-assessment component of a deployed agricultural decision-support system.

It is important to clarify that CRCE is introduced here as a study-specific prevalence-alignment measure rather than a replacement for conventional probabilistic calibration metrics. Its purpose is to quantify whether the aggregate frequency of Critical-risk assignments reflects the observed prevalence of the disease designated as requiring urgent intervention within the predefined business-context schema. Accordingly, CRCE evaluates population-level risk-assignment behavior and does not establish per-image correspondence or probability calibration. The near agreement observed for Gemma should therefore be interpreted as distributional alignment, whereas Qwen's larger deviation indicates systematic risk over-assignment. Individual-level validation remains an important direction for future work.

\subsubsection{Confusion-Matrix Analysis Across Datasets}
\label{sec:confusion_matrix_analysis}

\begin{figure*}[htbp]
    \centering
    \includegraphics[width=0.90\textwidth]
    {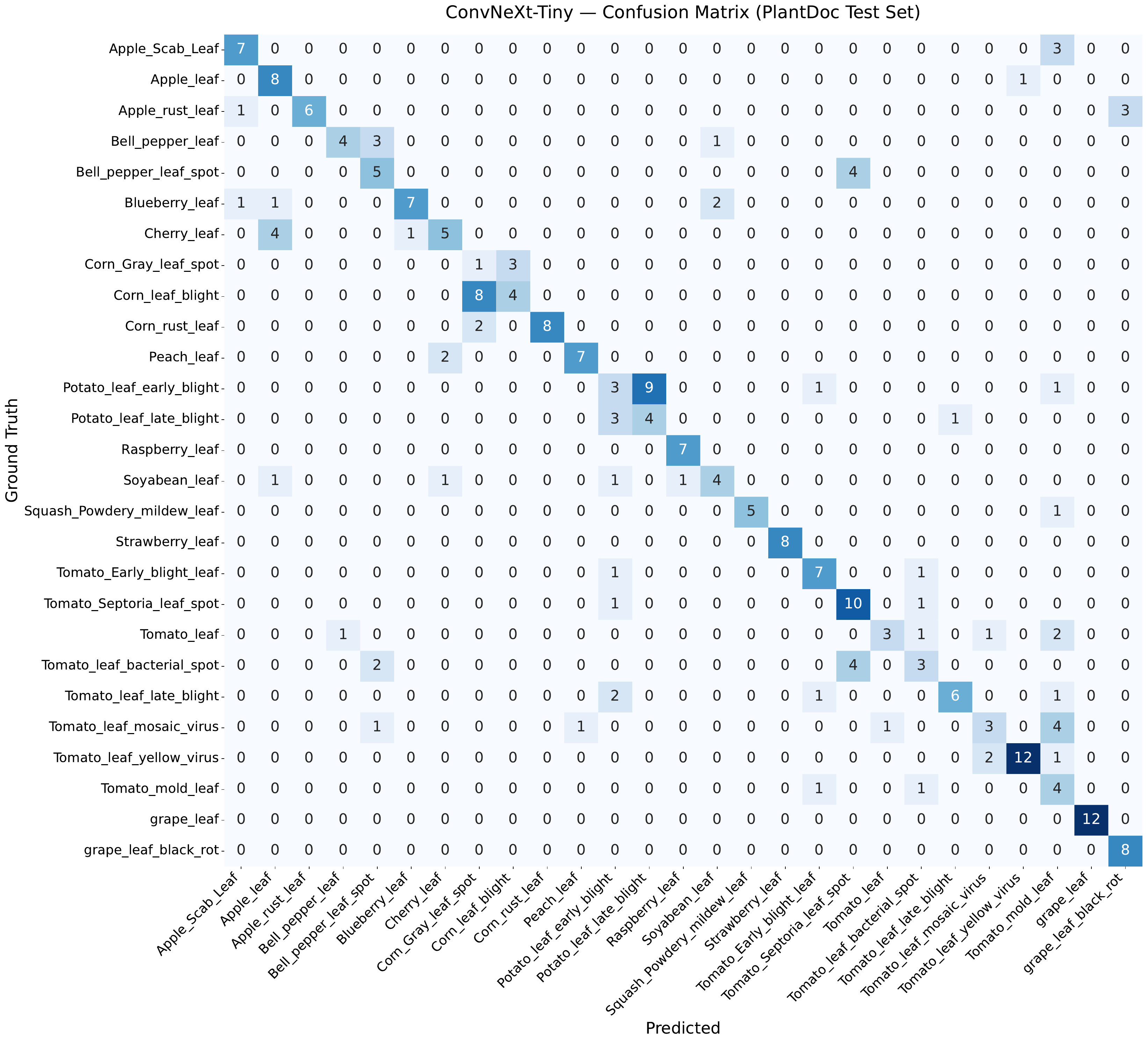}
    \caption{Confusion matrix of ConvNeXt-Tiny predictions on the 
    27-class PlantDoc test set. Rows represent ground-truth classes, 
    whereas columns represent predicted classes. Although most 
    predictions are concentrated along the main diagonal, off-diagonal 
    errors remain among visually similar disease categories, especially 
    within the tomato classes. The matrix illustrates the difficulty of 
    fine-grained disease recognition under heterogeneous backgrounds, 
    illumination conditions, symptom appearances, and limited 
    class-specific test support.}
    \label{fig:confusion_plantdoc}
\end{figure*}

\begin{figure*}[htbp]
    \centering
    \includegraphics[width=0.62\linewidth]
    {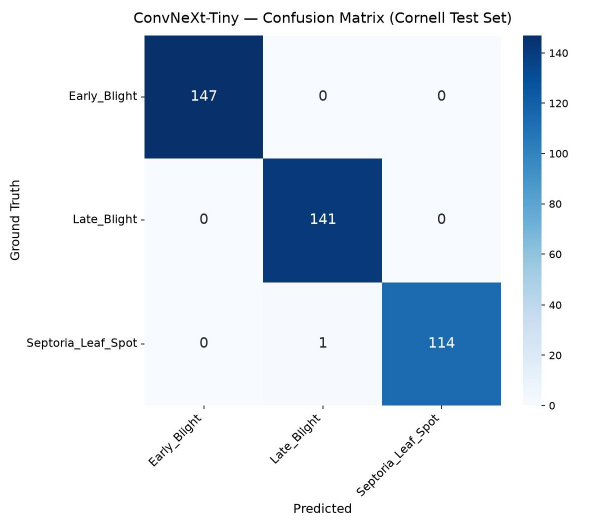}
    \caption{Confusion matrix of ConvNeXt-Tiny predictions on the 
    Cornell Stage~2 test set. Rows denote ground-truth disease classes 
    and columns denote predicted classes. The strongly diagonal 
    distribution demonstrates high discrimination among Early Blight, 
    Late Blight, and Septoria Leaf Spot under the Stage~2 field-acquisition 
    conditions, with only a very small number of cross-class errors.}
    \label{fig:confusion_cornell_stage2}
\end{figure*}

\begin{figure*}[htbp]
    \centering
    \includegraphics[width=0.62\linewidth]
    {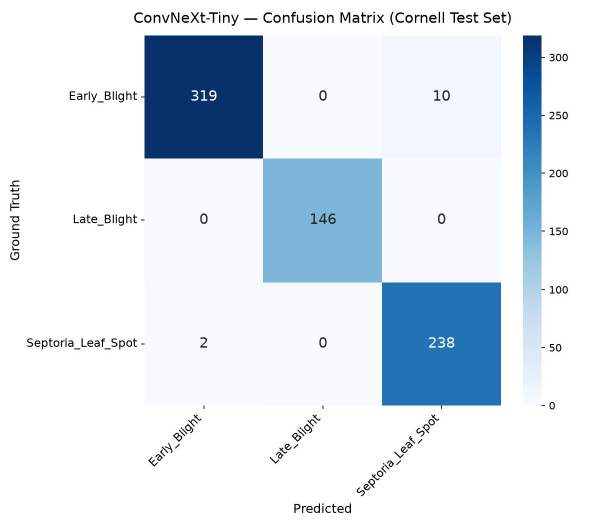}
    \caption{Confusion matrix of ConvNeXt-Tiny predictions on the 
    Cornell Stage~4 test set. The matrix remains strongly concentrated 
    along the main diagonal, indicating robust discrimination among the 
    three tomato foliar diseases. The principal residual confusion 
    occurs when a small number of Septoria Leaf Spot images are 
    classified as Early Blight, reflecting the partially overlapping 
    necrotic lesion characteristics of these diseases under natural 
    field conditions.}
    \label{fig:confusion_cornell_stage4}
\end{figure*}

Figures~\ref{fig:confusion_plantdoc}--\ref{fig:confusion_cornell_stage4} 
present the ConvNeXt-Tiny confusion matrices for the PlantDoc, Cornell 
Stage~2, and Cornell Stage~4 test sets, respectively. Separating the 
matrices improves the visibility of individual class-level errors, 
particularly for PlantDoc, which contains 27 classes and therefore 
requires substantially more display space than the two three-class 
Cornell datasets.

Figure~\ref{fig:confusion_plantdoc} shows that the errors on PlantDoc 
are distributed across multiple off-diagonal cells. Confusion occurs 
primarily among classes with similar visual symptoms or closely related 
crop--disease combinations, particularly within the tomato disease 
categories. This behavior is consistent with the class-wise results 
reported in Section~\ref{sec:results_plantdoc}. The relatively small 
number of test samples per class, substantial variations in image 
background, illumination, scale, and leaf orientation, and visual 
similarity among disease symptoms collectively make PlantDoc a 
challenging fine-grained classification task. Nevertheless, the 
dominant diagonal pattern indicates that ConvNeXt-Tiny retains useful 
class-discriminative capability across the heterogeneous 27-class 
benchmark.

In comparison, the Cornell Stage~2 confusion matrix in 
Fig.~\ref{fig:confusion_cornell_stage2} is strongly concentrated along 
the main diagonal. ConvNeXt-Tiny correctly distinguishes nearly all 
Early Blight, Late Blight, and Septoria Leaf Spot images, with only 
minimal cross-class confusion. This result agrees with the near-ceiling 
accuracy and class-wise precision, recall, and F1-score reported for 
Stage~2. However, the matrix should be interpreted in conjunction with 
the limited three-class taxonomy and the temporal similarity among 
continuously acquired field frames, both of which make direct comparison 
with the 27-class PlantDoc matrix inappropriate.

Figure~\ref{fig:confusion_cornell_stage4} presents the corresponding 
results for the larger Cornell Stage~4 test set. The matrix remains 
predominantly diagonal, demonstrating that ConvNeXt-Tiny maintains 
strong classification performance during the later and larger field 
acquisition campaign. The principal residual error involves a small 
number of Septoria Leaf Spot images classified as Early Blight. This 
confusion is biologically and visually plausible because both diseases 
can produce necrotic leaf lesions whose color, shape, and spatial 
distribution may appear similar under variable field illumination, 
particularly when symptoms are viewed at the canopy level. Despite 
these residual errors, Late Blight remains especially well separated 
from the other two disease classes.

Overall, the three confusion matrices demonstrate two distinct 
classification regimes. PlantDoc evaluates fine-grained recognition 
across 27 heterogeneous crop and disease classes and consequently 
exhibits broader interclass confusion. In contrast, the Cornell 
datasets evaluate a narrower three-disease tomato classification task 
and produce substantially stronger diagonal concentration. The Cornell 
results support the consistency of ConvNeXt-Tiny across two field 
acquisition campaigns, but their higher accuracies should not be 
attributed solely to improved robustness because class-set size, image 
acquisition modality, and temporal correlation also differ substantially 
from PlantDoc.

\subsubsection{Qualitative XAI Report Examples}

Beyond aggregate metrics, we illustrate the framework's qualitative reasoning on representative examples from each dataset.
For a PlantDoc Tomato Early Blight image where EfficientNet-B3 predicted Early Blight (72\%) and ConvNeXt-Tiny predicted Tomato Leaf (58\%), Gemma produced:
\begin{quote}
\small\textit{"final\_diagnosis": "Tomato\_Early\_blight\_leaf", "confidence": "Medium", "arbitration\_reasoning": "EfficientNet-B3 shows higher confidence (72\%) for Early Blight while ConvNeXt-Tiny's top prediction (Tomato\_leaf, 58\%) is not a disease class, suggesting it is misclassifying the background. The visual evidence of concentric necrotic rings confirms Early Blight.", "risk\_level": "High", "business\_recommendation": "Apply mancozeb-based fungicide within 48 hours. Estimated yield protection: €50,000.", "treatment\_window\_hours": 48}
\end{quote}

On the Cornell Stage~2 dataset, a genuinely conflicting case saw EfficientNet-B3 split between Late Blight (43.8\%) and Early Blight (36.8\%), while ConvNeXt-Tiny expressed overwhelming confidence (98.4\%) in Early Blight; Gemma's response reads:
\begin{quote}
\small\textit{"final\_diagnosis": "Early\_Blight", "confidence": "High", "arbitration\_reasoning": "The two models strongly disagreed. EfficientNet-B3 was split between Late and Early Blight, while ConvNeXt-Tiny provided overwhelming confidence (98.4\%) in Early Blight. The visual evidence, showing dark, somewhat circular lesions characteristic of early blight on the leaves and stems, aligns best with the ConvNeXt-Tiny prediction.", "risk\_level": "High", ...}
\end{quote}

On the Cornell Stage~4 dataset, a Septoria Leaf Spot image on which both CNNs agreed (68.0\% and 70.9\% confidence respectively) prompted Gemma to explicitly note the underlying business context:
\begin{quote}
\small\textit{"final\_diagnosis": "Septoria\_Leaf\_Spot", "confidence": "High", "arbitration\_reasoning": "Both models show very high agreement, with Septoria\_Leaf\_Spot being the leading prediction in both cases (68.0\% vs 70.9\%). The near-identical top predictions allow for a decisive classification.", "risk\_level": "Medium", "business\_recommendation": "Implement a prophylactic fungicide application targeting Septoria species within the next 48 hours to mitigate further spread, aiming to protect the €50,000 crop value.", "treatment\_window\_hours": 48}
\end{quote}

Across all three datasets, MLLM-generated reports consistently contain: (1)~a final diagnosis with stated confidence level; (2)~explicit arbitration reasoning referencing specific confidence scores from both CNN experts; (3)~a description of visual symptoms observed in the image; (4)~a risk level calibrated to crop value and disease urgency; (5)~a specific treatment recommendation; and (6)~a treatment window in hours---demonstrating that the explainability benefits of the H\textsuperscript{2}MAF framework are not an artifact of any single dataset but a consistent property of the semantic-fusion layer, replicated on genuine field imagery that the underlying MLLMs have never seen during pre-training.

\section{Discussion}
\label{sec:Discussion}

\subsection{Why MLLM Arbitration Value Tracks the Conflict Rate}

The cross-dataset synthesis in Section~\ref{sec:results_synthesis} shows that MLLM arbitration gain is not a fixed property of the framework but scales with how often the two CNN experts disagree; the results was found to be large and reliable on PlantDoc (41.7\% conflict, +7.6~points) and meaningful on Stage~4 (4.1\% conflict, +13.8~points on that smaller subset), whereas it was statistically evaluation on Stage~2 was not appropriate because of small dataset (1.7\% conflict, only 7 images).
We attribute this to three complementary mechanisms, all of which require a genuine disagreement signal to operate on.

\textbf{Semantic disambiguation.}
CNNs discriminate through texture pattern matching; they cannot reason about the semantic distinction between diseases with overlapping visual signatures (e.g., Septoria leaf spot vs.\ bacterial spot on PlantDoc, or Septoria vs.\ Early Blight on Stage~4).
MLLMs bring pre-trained knowledge about symptom semantics, allowing them to leverage clues, lesion margin sharpness, halo colouration, distribution across the leaf surface that pure texture features cannot encode.

\textbf{Confidence-weighted arbitration.}
When the two CNNs disagree, the MLLM receives both sets of confidence scores alongside the image.
In most PlantDoc conflict cases, it correctly defers to the higher-confidence expert (80 cases of ``agrees with ConvNeXt-Tiny only''), achieving 50\% accuracy---8~points above EfficientNet-B3's 32.4\% on the same images; the qualitative Stage~2 example in Section~\ref{sec:results_synthesis} shows the identical mechanism operating on real field imagery, where Gemma explicitly cites ConvNeXt-Tiny's 98.4\% confidence as the deciding factor.

\textbf{Visual grounding on low-confidence images.}
Table~\ref{tab:confidence} confirms that on PlantDoc images where both CNNs express low confidence ($<$0.5), Gemma achieves 41.7\% accuracy vs.\ 35.4\% for the CNN alone a 6.3-point gain from independent visual inspection.
The Cornell datasets, by contrast, contain almost no low-confidence images (0 in both test sets under the same 0.5 threshold), which is itself part of why raw arbitration opportunity is scarcer on real, field-curated imagery of a narrower disease taxonomy.

\subsection{The Unconstrained Override Problem is a General Design Boundary, Not a PlantDoc Artifact}

The most cautionary finding of this study, first observed on PlantDoc dataset (13.6--17.6\% accuracy when the MLLM overrides both CNNs), \emph{replicates and in some respects worsens} on the larger, more diverse Stage~4 dataset, where Qwen's override rate rises to 7.6\% of the test set (54/715 images) with only 1.9\% accuracy on those cases directly explaining Qwen's lower aggregate Stage~4 accuracy relative to Stage~2. This cross-dataset replication substantially strengthens the case that unconstrained override is a structural property of current MLLMs used as arbitrators, not an artifact of any single dataset's class distribution or image style. Prompt engineering alone cannot fully suppress this behaviour: MLLMs are stochastic, and rare hallucinations produce plausible-sounding but incorrect diagnoses, particularly (as the Stage~4 result suggests) as the volume and diversity of unfamiliar field imagery increases.

This finding motivates a practical design principle for multi-agent agricultural information-fusion systems: \emph{constrain MLLM autonomy to conflict resolution between pre-computed signals}; do not allow the MLLM to generate diagnoses from scratch. Future work should explore ``veto constraints'' that force the MLLM's final diagnosis to belong to the union of top-$K$ predictions from both CNNs, a constraint that based on the override-collapse statistics reported here, would likely improve aggregate accuracy on every dataset in this study, most dramatically for Qwen on Stage~4.

\subsection{Model Personalities and Risk Calibration: A Deployment-Critical Finding}

The risk-prevalence calibration analysis in Section~\ref{sec:results_synthesis} is, to our knowledge, a novel contribution not present in prior MLLM-for-agriculture literature: rather than merely observing that two models assign different risk distributions, we quantify calibration error against an objective, dataset-derived ground truth (true prevalence of the sole ``high risk-aversion'' disease class).

The result Gemma's CRCE of 0.14 to 0.5~pp versus Qwen's 3.5--14.4~pp, replicated on two independent real-world datasets quantifies what was previously only a qualitative observation in Section~\ref{sec:results_synthesis}, namely that Qwen assigns substantially more Critical-risk labels than Gemma (the 3.7$\times$ gap first reported on PlantDoc in Table~\ref{tab:risk}).

This asymmetry has direct operational implications.
A system deployed with Qwen's risk model would trigger unnecessary and costly interventions well beyond the true rate of urgent disease, a bias that grows rather than shrinks as the deployment dataset grows (3.5~points on Stage~2 vs.\ 14.4~points on Stage~4). Gemma's near-perfect tracking of true prevalence on both Cornell datasets suggests it has either better internalised epidemiological base rates during pre-training or applies a more conservative default risk assignment that happens to align with this domain; distinguishing between these explanations, and testing whether the calibration property generalises to other disease taxonomies with different true prevalence rates, is an important direction for future work.
Meanwhile, we recommend that any production deployment of an MLLM-based agricultural risk-assessment layer include a calibration-verification step comparable to the one presented in Section~\ref{sec:results_synthesis}, using held-out data with known disease prevalence, before the model's risk labels are used to trigger real interventions.
Candidate post-hoc calibration methods include temperature scaling and Platt scaling, both of which learn a single calibration parameter on a held-out validation set to align a model's output distribution with observed ground truth; in the present context, such methods could be applied to the MLLM's risk-level assignment (rather than its classification logits) to systematically shift Qwen's over-flagged Critical-risk rate toward the true epidemiological base rate.

\subsection{Practical Deployment Considerations}

Deploying the H\textsuperscript{2}MAF framework in production agricultural settings introduces several engineering and operational considerations that extend beyond the laboratory evaluation presented here.

\textbf{Latency requirements.}
For real-time field scouting applications, total pipeline latency (CNN inference + JSON assembly + MLLM inference) must be compatible with operator workflows.

On the H100 hardware used in this study, CNN inference for a single image takes less than 50~ms per model; MLLM inference for a single image with a 512-token prompt takes approximately 8--15~seconds depending on response length, consistently across all three datasets. For asynchronous applications (nightly batch reports, post-harvest analysis), this latency is acceptable. For real-time scouting drones or automated greenhouse sorting lines that inspect individual plants at high throughput, deployment scenarios analogous to the continuous-capture Cornell datasets, a smaller MLLM (2B parameters) or a hardware-optimised quantised variant (GGUF Q4\_K\_M) would be required to achieve sub-second latency compatible with the platform's native $\sim$0.3--1~second frame rate.

\textbf{Edge deployment.}
Both Gemma~4~E4B (4.5B effective parameters) and Qwen3.5~4B (4B parameters) are specifically designed for on-device deployment; Google DeepMind \cite{Gemma4_2025} targets edge devices including laptops and high-end mobile hardware.
With 4-bit quantisation, Gemma~4~E4B requires approximately 5--6~GB VRAM, fitting within the footprint of NVIDIA Jetson AGX Orin (64~GB shared memory) or equivalent agricultural IoT hardware---precisely the class of embedded compute available on an autonomous field-imaging platform such as the one that produced the Cornell datasets.

This compact memory footprint makes the full pipeline deployable at the farm edge without cloud connectivity, which is critical for regions with unreliable internet infrastructure.

\textbf{Model update and maintenance.}
A key advantage of the modular architecture is independent updatability of each component.
The two CNN perceptual experts can be retrained as new disease strains emerge or as new field-capture campaigns (such as the Stage~2$\to$Stage~4 progression demonstrated in this study) become available, without modifying the MLLM layer or the JSON artifact schema.
Conversely, if a stronger MLLM is released, it can be swapped in without CNN retraining, provided the prompt format remains compatible.

This modular decoupling contrasts with end-to-end MLLM approaches \cite{roumeliotis11373381}, where the entire model must be retrained to incorporate new disease class information.

\subsection{Economic and Agronomic Impact Analysis}

The business-intelligence translation performed by the MLLM layer transforms a classification result into a recommendation for economically quantified intervention, on both benchmark and real-world data.

Using Gemma's PlantDoc risk distribution (Table~\ref{tab:risk}), the framework assigns 25 images to Critical-risk and 99 to High-risk, generating 124 actionable intervention recommendations---each accompanied by a crop-value-at-risk figure drawn from the static business-context lookup table defined in Section~\ref{sec:Methodology} (e.g., \euro{}50,000 for tomato), of which 68.5\% are associated with correct diagnoses (based on test accuracy).

On the Cornell datasets, the near-ceiling classification accuracy (96.8--99.8\%) means the corresponding proportion of correctly-diagnosed recommendations is substantially higher (91.8--99.8\%), suggesting that as classification accuracy improves with better real-world training data, the reliability of the downstream business-intelligence layer improves commensurately an encouraging signal for eventual field deployment, tempered by the session-correlation caveat discussed in Section~\ref{sec:dataset_cornell}.

The 13.9\% damage rate observed for Gemma on PlantDoc (Table~\ref{tab:benefit}), and the 6.4\% damage rate observed for Qwen on Stage~4 the highest damage rate for any model on any dataset in this study both imply that a fraction of recommendations are based on an incorrect diagnosis.

For Critical-risk cases, an incorrect recommendation (e.g., applying fungicide for the wrong pathogen) wastes treatment cost but carries comparatively low risk of harm.
For cases where a diseased plant is misclassified as healthy or Low-risk, no intervention is triggered the more operationally dangerous failure mode, and one that our per-class analysis (Sections~\ref{sec:results_plantdoc}--\ref{sec:results_stage4}) suggests is rare but non-zero on every dataset studied.

Calibrating risk thresholds to match local epidemiological priors and crop-specific economic injury levels is therefore a critical step before production deployment, and we recommend treating MLLM risk outputs as soft scores subject to post-hoc recalibration, using the CRCE-style ground-truth comparison introduced in Section~\ref{sec:results_synthesis} rather than hard decision boundaries.

Recent state-of-the-art plant disease recognition methods have increasingly adopted vision transformers\cite{ali2025plant, ouamane2025optimized, zhang2025convolutional}, CNN-transformer hybrids \cite{mondal2026hybrid, jia2025convtransnet}, attention-enhanced architectures \cite{haque2026attention, duhan2024investigating, gonzalez2025enhancing}, and probability-level ensembles to improve classification under complex imaging conditions \cite{srinivasu2026deep, korkmaz2025explainable}. Although these methods frequently achieve strong benchmark accuracy, most remain confined to label prediction and provide limited support for reconciling contradictory model evidence or translating predictions into operational decisions \cite{anwar2026empowering, goluguri2026revolutionizing}. Emerging MLLM-based approaches offer visual question answering and natural-language disease interpretation \cite{shuai2026agrimapo, saha2025fusing, zhu2025potato}; however, direct end-to-end diagnosis may introduce hallucinated classes, unsupported treatments, and poorly calibrated confidence. Conventional XAI techniques, including Grad-CAM\cite{selvaraju2016grad}, SHAP \cite{lundberg2017unified}, and LIME\cite{ribeiro2016should}, improve visual interpretability but generally require expert interpretation and do not perform semantic conflict arbitration \cite{nguyen2021evaluation, nanuvala2026comparative, akgundougdu2025explainable}.

Our H\textsuperscript{2}MAF proposed in this study addresses these limitations by preserving specialist CNNs as the primary perceptual layer, constraining MLLM reasoning through a structured JSON evidence artefact, and explicitly analyzing agreement, disagreement, override behavior, and downstream risk assignment. Its evaluation across PlantDoc and two robot-acquired Cornell datasets further extends validation beyond curated benchmark imagery. Nevertheless, the framework should be interpreted as an initial step toward deployable multimodel decision support. The Cornell experiments use a narrower three-class taxonomy, continuously acquired frames may contain temporal correlation, and the zero-shot MLLMs rely on static business-context records rather than live epidemiological, weather, or market information. Future comparisons should therefore include confidence-weighted ensembles, transformer classifiers, domain-adapted MLLMs, retrieval-augmented agronomic reasoning, and session-independent field evaluation to establish whether semantic arbitration remains beneficial across farms, crops, disease prevalence levels, and acquisition systems.

\subsection{Limitations}

\textbf{Test set size and per-class variance.}
The PlantDoc test set contains only 252 images (4--15 per class), yielding high per-class metric variance; on Cornell Stage~2, only 7 of 403 test images produced disagreement between the two CNN experts, and the resulting negative apparent MLLM gain on those 7 images (Section~\ref{sec:results_synthesis}) should be treated as inconclusive rather than as evidence of framework failure.
Conclusions about specific classes or small subsets should be treated as indicative rather than definitive.

\textbf{Temporal correlation and image-format heterogeneity in the Cornell datasets.}
As discussed in Section~\ref{sec:dataset_cornell}, both Cornell datasets consist of continuous video-frame captures, and our frame-level stratified random split may place near-duplicate frames of the same physical lesion into different partitions of the split.
% This temporal leakage almost certainly contributes to the very high (96--99.8\%) CNN accuracy observed on these datasets relative to PlantDoc, and we recommend that future work using this data modality adopt session-level (rather than frame-level) train/validation/test splitting to obtain a more conservative and realistic accuracy estimate.
This temporal leakage is a known source of accuracy inflation in video-derived datasets and contributes to the very high CNN accuracy observed on the Cornell datasets (96--99.8\%) compared with the substantially lower accuracy achieved on PlantDoc (59--64\%), and we recommend that future work using this data modality adopt session-level (rather than frame-level) train/validation/test splitting to obtain a more conservative and realistic accuracy estimate.
Additionally, the Cornell images are full plant-canopy views captured at a consistent distance by the robot platform, whereas PlantDoc images are individual leaf photographs with highly variable framing, background clutter, and scale.
Future work should explore localised lesion-patch cropping to extract individual symptomatic regions from the canopy-level images, which would both improve consistency with the PlantDoc image format and reduce the background-to-lesion ratio that may currently inflate classification confidence.

This limitation is essential for correctly interpreting the magnitude---though not the direction or generalisability-of the cross-dataset findings in Section~\ref{sec:results_synthesis}, which rely on relative quantities (conflict rate, calibration error) that are considerably more robust to this effect than raw accuracy.

\textbf{Zero-shot MLLM inference.}
Both MLLMs are used in zero-shot mode on every dataset.
Domain-specific fine-tuning (e.g., on agricultural XAI reports, or specifically on Cornell field imagery) would likely improve both accuracy and risk calibration, and represents a natural next step now that two real-world training corpora are available.

\textbf{Business context placeholder.}
The current implementation uses static crop-value and risk-aversion lookup tables for both PlantDoc and the Cornell datasets.
A production system should integrate live market prices, field-sensor weather data, and historical disease pressure records specific to the deployment region.

% \textbf{Dataset access.}
% The two Cornell datasets used in this study are closed, non-public data shared under a research collaboration and are not redistributed as part of this paper's public code release (Section~\ref{sec:Conclusion}); this is an inherent trade-off of using genuine, non-curated field data rather than a public benchmark, and we encourage other groups with access to comparable proprietary field-imaging data to replicate the cross-dataset methodology introduced here.

%% ─────────────────────────────────────────────────────────────────────────────
\section{Conclusion}
\label{sec:Conclusion}

This paper introduced the Hybrid Hierarchical Multi-Agent Framework (H\textsuperscript{2}MAF), a two-stage information-fusion pipeline that first fuses two architecturally distinct CNN perceptual experts at the decision level and then fuses their output with pre-trained multimodal-LLM semantic knowledge at the reasoning level, translating pixel-level classification into a structured, actionable Explainable AI report.
We validated this framework across \textbf{three independent datasets totalling 14,364 images and 1,370 held-out test images}: the internet-curated PlantDoc benchmark, and two previously unpublished, continuously-captured, robot-acquired real-world field datasets generated by the co-authors (Stage~2, 20~GB; Stage~4, 40~GB)to our knowledge the first evaluation of an MLLM-based conflict-arbitration framework on context-specific, highly realisitic non-public agricultural field data. Among the four acquisition stages, Stage~2 and Stage~4 were selected for the present study to represent two temporally separated points in disease progression while providing sufficiently large image sets for model training and evaluation.

On PlantDoc, MLLM arbitration improved top-1 accuracy from 63.9\% (best CNN) to 68.5\% (Gemma~4~E4B), concentrated on the 41.7\% of images where the two CNN experts disagreed (+7.6~points over the stronger CNN).
On the Cornell datasets, both CNN experts already achieve 96.8--99.8\% accuracy and agree on 95.9--98.3\% of images, and the cross-dataset synthesis established that MLLM arbitration gain scales with this conflict rate: substantial and reliable when the conflict subset is sufficiently large for statistical inference (PlantDoc, $n=105$; Stage~4, $n=29$), and statistically inconclusive when the conflict subset is too small to support reliable estimation (Stage~2, $n=7$).
Across all three datasets, unconstrained MLLM override of both CNN experts collapsed accuracy to 0--17.6\%, replicating and strengthening a design principle first observed on a single dataset alone: MLLMs should resolve conflicts between CNN signals, not generate diagnoses de novo.

A novel risk-prevalence calibration analysis, replicated on both Cornell datasets against an objective ground truth (true Late Blight prevalence), showed that Gemma's assigned Critical-risk rate tracks true urgent-disease prevalence to within 0.14--0.5\%, while Qwen systematically over-flags by 3.5--14.4 points, a consistent, quantifiable, model-specific risk-assessment bias with direct implications for which MLLM should be selected as the risk-assessment component of a deployed agricultural decision-support system.

Future work will explore: (i)~domain-specific MLLM fine-tuning on curated agricultural XAI corpora drawn from the Cornell field data itself; (ii)~constrained decoding that restricts the MLLM diagnosis space to the union of top-$K$ CNN predictions, directly targeting the override-collapse failure mode; (iii)~session-level (rather than frame-level) splitting of continuously-captured field video to obtain more conservative, deployment-realistic accuracy estimates; (iv)~extension to additional real-world field campaigns beyond Stage~2 and Stage~4, to further test the generalizability of the conflict-dependent utility and risk-calibration findings reported here; and (v)~systematic risk calibration using temperature or Platt scaling to align MLLM risk priors, particularly Qwen's, with deployment-specific epidemiological base rates.

\noindent The PlantDoc-based code, trained models, and evaluation artefacts are publicly available at: \url{https://github.com/Applied-AI-Research-Lab/Explainable-AI-Plant-Disease-Detection} \cite{ExplainableAIPlantDiseaseDetection}. The two Cornell field datasets are closed, non-public data shared under research collaboration with Cornell University's Automation and Robotics Laboratory and are not included in this release; the training, artefact-generation, and evaluation code used to process them is otherwise identical to the public PlantDoc pipeline and is likewise available in the repository above.
\section*{Author Contributions}
\textbf{Ranjan Sapkota:} Conceptualization, Methodology, Software, Validation, Formal analysis, Investigation, Data curation, Writing – original draft, Writing – review and editing, Visualization, Project administration. \textbf{Konstantinos I. Roumeliotis:} Conceptualization, Methodology, Software, Validation, Formal analysis, Investigation, Data curation, Writing – original draft, Writing – review and editing, Visualization, Project administration. \textbf{Pengyao Xie:} Data Acquisition, Formal analysis, Writing – review and editing  \textbf{Nikolaos D. Tselikas:} Conceptualization, Validation, Resources, Writing – review and editing, Supervision. \textbf{Lirong Xiang:} Data Acquisition, Writing – review and editing, Supervision, Funding acquisition. \textbf{Manoj Karkee:} Conceptualization, Resources, Writing – review and editing, Supervision, Funding acquisition.

\section*{Acknowledgement} This work is supported by the National Science Foundation (NSF) and the United States Department of Agriculture (USDA), National Institute of Food and Agriculture (NIFA), through the “Artificial Intelligence (AI) Institute for Agriculture” program under Award Numbers AWD003473 and AWD004595, and USDA-NIFA Accession Number 1029004 for the project titled “Robotic Blossom Thinning with Soft Manipulators.” Additional support was provided through USDA-NIFA Grant Number 2024-67022-41788, Accession Number 1031712, under the project “Expanding UCF AI Research To Novel Agricultural Engineering Applications (PARTNER).” and USDA-NIFA 2024-67021-42788 under the project “PARTNERSHIP: High-Throughput Multi-Scale Sensing for Tomato Disease Phenotyping under Field Conditions.” 
Additionally, this research utilised equipment that was acquired through funding by the European Union – NextGenerationEU and the Recovery and Resilience Facility (Greece 2.0), under the project SUB2: “Universities of Excellence” (Project code: OPS TA 5180665). All experiments were conducted on a server provided by the University of the Peloponnese, equipped with four NVIDIA H100~NVL GPUs (96~GB VRAM each) and an Intel Xeon Platinum~8452Y processor (36 cores).

\section*{Declarations}
The authors declare no conflicts of interest.

\section*{Declaration of generative AI and AI-assisted technologies in the manuscript preparation process}
During the preparation of this work the author(s) used ChatGPT and Grammarly in order to enhance grammatical accuracy and refine sentence structure. After using this tool/service, the author(s) reviewed and edited the content as needed and take(s) full responsibility for the content of the published article.

\section*{Data Availability}
All training code, CNN architectures, JSON artefact-generation scripts, MLLM inference scripts, and evaluation code developed in this study are publicly available on GitHub as open-source under the Apache-2.0 license : https://github.com/Applied-AI-Research-Lab/Explainable-AI-Plant-Disease-Detection. 

The PlantDoc \cite{Singh_2020,GuerreroSerna2025PlantDoc} dataset is publicly accessible through their respective repositories.

The two Cornell real-world field datasets (Stage~2 and Stage~4) analysed in this study are closed, non-public data shared by the Automation and Robotics Laboratory, Cornell University, under a research collaboration agreement. The datasets are available from the authors upon reasonable request. To support transparency and reproducibility, the complete processing pipeline used to analyze these datasets is publicly available in the code repository referenced above and can be applied to comparable field-imaging datasets. 

\bibliographystyle{unsrtnat}

\bibliography{references}

\appendix
\newpage

% \bibliographystyle{elsarticle-harv} 
% \bibliography{example}

\printcredits




\end{document}